\pdfoutput=1
\documentclass[11pt]{article}

\usepackage{ACL2023}

\usepackage{times}
\usepackage{latexsym}
\usepackage{amsmath}
\usepackage{graphicx}
\usepackage{subfig}
\usepackage{xcolor,colortbl}

\usepackage[T1]{fontenc}
\usepackage[utf8]{inputenc}

\usepackage{microtype}

\usepackage{inconsolata}
\usepackage{multirow}
\title{Exploring the Relationship between Alignment and Cross-lingual Transfer in Multilingual Transformers}

\author{Félix Gaschi \textsuperscript{\rm 1,2}, Patricio Cerda \textsuperscript{\rm 1}, Parisa Rastin  \textsuperscript{\rm 2}, Yannick Toussaint \textsuperscript{\rm 2}\\
 \textsuperscript{\rm 1}Posos,  \textsuperscript{\rm 2}LORIA \\
  \texttt{\{felix.gaschi,parisa.rastin,yannick.toussaint\}@loria.fr} \\
  \texttt{patricio@posos.fr} 
}

\begin{document}
\maketitle

\begin{abstract}
Without any explicit cross-lingual training data, multilingual language models can achieve cross-lingual transfer. One common way to improve this transfer is to perform realignment steps before fine-tuning, i.e., to train the model to build similar representations for pairs of words from translated sentences. But such realignment methods were found to not always improve results across languages and tasks, which raises the question of whether aligned representations are truly beneficial for cross-lingual transfer. We provide evidence that alignment is actually significantly correlated with cross-lingual transfer across languages, models and random seeds. We show that fine-tuning can have a significant impact on alignment, depending mainly on the downstream task and the model. Finally, we show that realignment can, in some instances, improve cross-lingual transfer, and we identify conditions in which realignment methods provide significant improvements. Namely, we find that realignment works better on tasks for which alignment is correlated with cross-lingual transfer when generalizing to a distant language and with smaller models, as well as when using a bilingual dictionary rather than FastAlign to extract realignment pairs. For example, for POS-tagging, between English and Arabic, realignment can bring a +15.8 accuracy improvement on distilmBERT, even outperforming XLM-R Large by 1.7. We thus advocate for further research on realignment methods for smaller multilingual models as an alternative to scaling.\end{abstract}

\section{Introduction}

\begin{figure}[t]
    \centering
    \includegraphics[width=\linewidth]{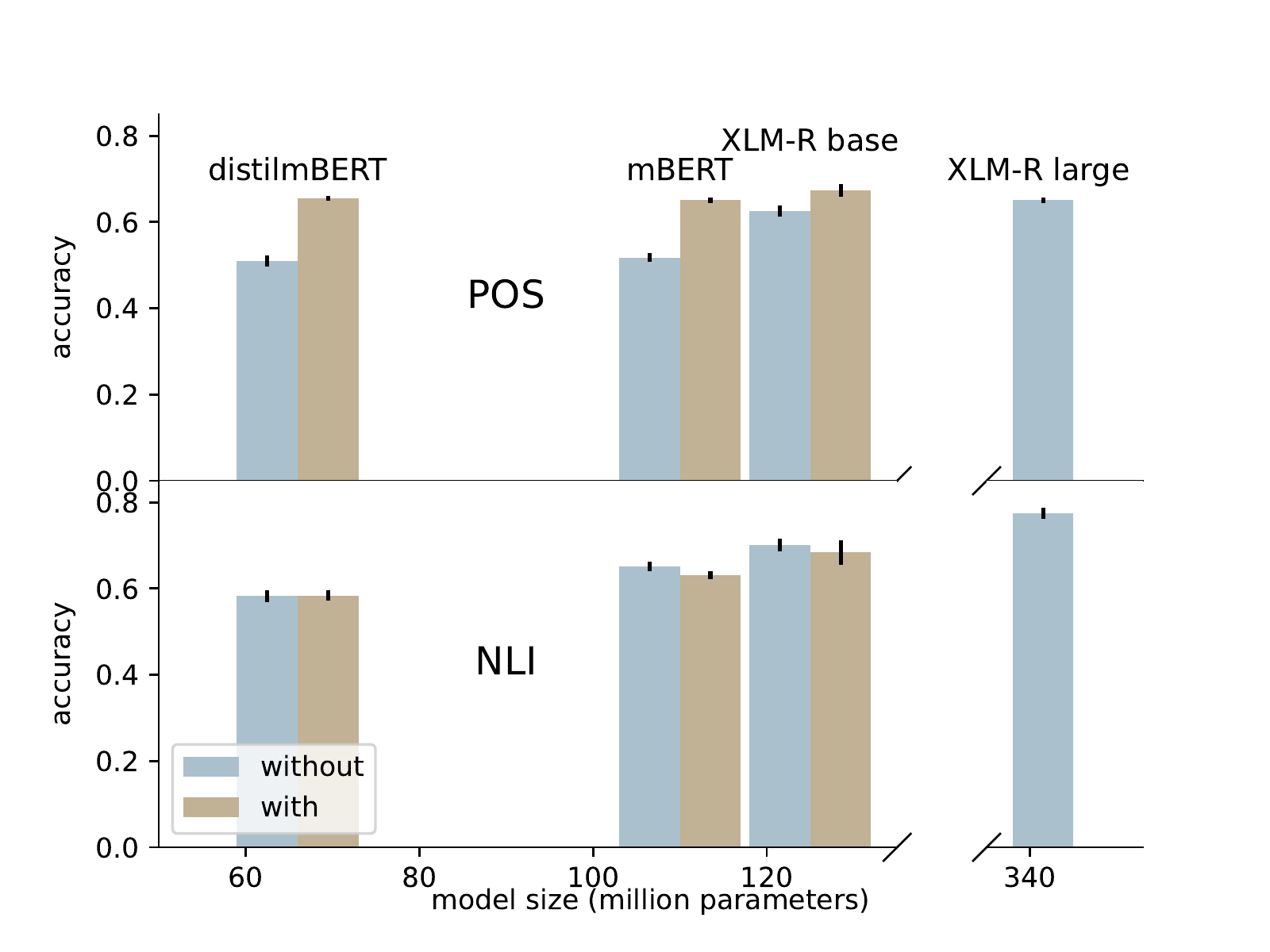}
    \caption{Cross-lingual transfer between English and Arabic with and without realignment, using a bilingual dictionary. For some tasks, realignment can make small models competitive with a large baseline.}
    \label{fig:summary}
\end{figure}

% TLDR of our contribution
With the more general aim of improving the understanding of Multilingual Large Language Models (MLLM), we study the link between the multilingual alignment of their representations and their ability to perform cross-lingual transfer learning, and investigate conditions for realignment methods to improve cross-lingual transfer.

% Explaining and citing Multilingual models, Cross-lingual Transfer succinctly
MLLMs, like mBERT \cite{devlin-etal-2019-bert} and XLM-R \cite{conneau-etal-2020-unsupervised}, are Transformer encoders \cite{vaswani-etal-2017} which show an effective ability to perform Cross-lingual Transfer Learning (CTL). Despite the absence of any explicit cross-lingual training signal, mBERT and XLM-R can be fine-tuned on a specific task in one language and then provide high accuracy when evaluated on another language on the same task \cite{pires-etal-2019-multilingual,wu-dredze-2019-beto}. By alleviating the need for training data for a specific task in all languages and for translation data which more than often lacks for non-English languages, CTL with MLLMs could help bridge the gap in NLP between English and other languages.

% CTL is not perfect, aka why do we want to improve CTL 
But the ability of MLLMs to generalize across languages is highly correlated with the similarity between the training language (often English) and the language to which we hope to transfer knowledge \cite{pires-etal-2019-multilingual,wu-dredze-2019-beto}. For distant and low-resources languages, CTL with mBERT can give worse results than fine-tuning a Transformer from scratch \cite{wu-dredze-2020-languages}.

% Introducing realignment methods
Realignment methods \cite{wu-dredze-2020-explicit}, sometimes called adjustment or explicit alignment, aim to improve the cross-lingual properties of an MLLM by trying to make similar words from different languages have closer representations. Realignment methods typically require a translation dataset and an alignment tool, like FastAlign \cite{dyer-etal-2013-simple}, to extract contextualized pairs of translated words that will be realigned.

% Failure of realignment methods
Despite some encouraging results on specific tasks, current realignment methods might not consistently improve cross-lingual zero-shot abilities of mBERT and XLM-R \cite{wu-dredze-2020-explicit}. When tested with several seeds on various fine-tuning tasks, improvements brought by realignment are not always significant and do not compare with the gain brought by scaling the model, e.g., from XLM-R Base to XLM-R Large. However, these realignment methods were not tried on smaller models like distilmBERT as we do here.

% NEW: is alignment linked with zero-shot transfer
The mitigated results of realignment methods raise the question of whether cross-lingual transfer is at all linked with multilingual alignment. If improving alignment does not necessarily improve CTL, then the two might not be correlated. Despite the ability of mBERT and XLM-R to perform CTL, there lacks consensus on whether they actually hold aligned representations \cite{gaschi-etal-2022}.

We thus investigate the link between alignment and CTL, with three contributions: (1) We find a high correlation between multilingual alignment and cross-lingual transfer for multilingual Transformers, (2) we show that, depending on the downstream task, fine-tuning on English can harm the alignment to different degrees, potentially harming cross-lingual transfer, and (3) we link our findings to realignment methods and identify conditions under which they seem to bring the most significant improvements to zero-shot transfer learning, particularly on smaller models as shown on Fig. \ref{fig:summary}.

\section{Related Work}

% Enumerating and differentiating different existing realignment methods
Current realignment methods are applied on a pre-trained model before fine-tuning in one language (typically English). Common tasks are Natural Language Inference (NLI), Named Entity Recognition (NER), Part-of-speech tagging (POS-tagging) or Question Answering (QA). The model is then expected to generalize better to other languages for the task than without the realignment. Realignment methods rely on pairs of words extracted from translated sentences using a word alignment tool, usually FastAlign \cite{dyer-etal-2013-simple}, but other tools like AWESOME-align \cite{dou-neubig-2021-word} could be used. Various realignment objectives are used to bring closer together the contextualized embeddings of words in such pairs: using a linear mapping \cite{wang-etal-2019-cross}, a $\ell_2$-based loss with regularization to avoid degenerative solutions \cite{Cao2020Multilingual,zhao-etal-2021-inducing}, or a contrastive loss \cite{pan-etal-2021-multilingual,wu-dredze-2020-explicit}.

%NOTE: talk about alignment data only in the Method
%NOTE: talk about loss more in details in the Method
%NOTE: talk about details of measuring alignment in Method

% Instances where realignment methods were shown to be limited
Existing realignment methods might not significantly improve cross-lingual transfer. Despite improvements on NLI \cite{Cao2020Multilingual,zhao-etal-2021-inducing,pan-etal-2021-multilingual} or on dependency parsing \cite{wang-etal-2019-cross}, the results might not hold across tasks and languages. A comparative study by \citet{kulshreshtha-etal-2020-cross} showed that methods based on linear mapping are effective only on "moderately close languages", whereas $\ell_2$-based loss improves results for "extremely distant languages". This latter $\ell_2$-loss was shown to work well on a NLI task, but not for all languages on a NER task, and to be even detrimental for QA tasks \cite{eftimov-etal-2022-impact}. Finally, \citet{wu-dredze-2020-explicit} compared linear mapping realignment, $\ell_2$-based realignment and contrastive learning on several tasks, languages and models, performing several runs. They found that existing methods do not bring consistent improvements over no realignment. 

Expecting realignment methods to succeed implies a direct link between the multilingual alignment of the representations produced by a model and its ability to perform CTL. However, there isn't any strong consensus on whether multilingual Transformers have well-aligned representations \cite{gaschi-etal-2022}, let alone on whether better-aligned representations lead to better CTL. 

Assessing the multilingual alignment of contextualized representations can take many forms. Pairs of words are extracted from translated sentences, usually with FastAlign or a bilingual dictionary \cite{gaschi-etal-2022}. Then, after building contextualized representations of the words of each pair, the distribution of their similarity can be compared  with that of random pairs of words \cite{Cao2020Multilingual}. But this method can lead to incorrect conclusions \cite{eftimov-etal-2022-impact}. A high overlap in the distribution of similarities between related and random pairs means that sometimes random pairs can have higher similarities than related pairs. But since those pairs do not necessarily involve the same words, a high overlap does not mean that any word is closer to an unrelated one than to a related one. An alternative is to compare a related pair to its neighbors \cite{eftimov-etal-2022-impact}, which shows that realignment methods indeed improve multilingual alignment and that fine-tuning can harm this alignment. Another similar approach consists in designing a nearest-neighbor search criterion. This was done for sentence-level representations \cite{pires-etal-2019-multilingual} and for word-level alignment \cite{conneau-etal-2020-emerging,gaschi-etal-2022}, showing that MLLMs like mBERT have a multilingual alignment that is competitive with static embeddings \cite{bojanowski-etal-2017-enriching} explicitly aligned with a supervised alignment method \cite{joulin-etal-2018-loss}.\looseness=-1
 
\section{Method}

To study the link between multilingual alignment and cross-lingual transfer (CTL), we need a way to evaluate alignment and CTL. We use a relative difference to evaluate CTL, we discuss different methods for evaluating alignment, and describe the realignment method used in our experiments.

\subsection{Evaluating cross-lingual transfer\label{section:evaluate_transfer}}

A model has high CTL abilities when, after fine-tuning for one language, it can obtain a high evaluation score on other languages. To evaluate it for a given task, we compute the relative difference between the evaluation metric $m_{\text{en}}$ on the English development set and the evaluation metric $m_{\text{tgt}}$ on the target language:
\begin{equation}\label{eq:transfer}
    \text{cross-lingual transfer} = \frac{m_{\text{tgt}} - m_{\text{en}}}{m_{\text{en}}}
\end{equation}The monolingual metric is a score between 0 and 1, like accuracy or f1-score, where higher is better. Then our metric gives scores between -1 and $+\infty$. A negative score is obtained if and only if $m_{\text{tgt}} < m_{\text{en}}$, which should always be the case in practice. Values closer to 0 then indicate better CTL for a specific task and language. 

It must be noted that for datasets where the target language test set is a translation of the English one, the normalization in Equation \ref{eq:transfer} allows the metric to boild down roughly to minus the proportion of correct answers in English that were misclassified when translated, assuming there isn't not to many misclassified English examples that were correctly classified in the target language, which should be the case since there are not that much misclassified English examples in general. 

\subsection{Evaluating alignment}\label{section:evaluation}

To evaluate multilingual alignment, we use the same method for extracting pairs of translated words with their context as \citet{gaschi-etal-2022}. Provided a source of related pairs of words from both languages, a fixed number of pairs of words are randomly selected and a nearest-neighbor search with cosine similarity is performed. The top-1 accuracy of the nearest-neighbor search is the score of the alignment evaluation.

To extract contextualized pairs of translated words from a translation dataset, FastAlign is the most widely used word aligner in realignment methods \cite{wu-dredze-2020-explicit,Cao2020Multilingual,zhao-etal-2021-inducing,wang-etal-2019-cross}, but it is prone to errors and thus generates noisy training realignment data \cite{pan-etal-2021-multilingual,gaschi-etal-2022}. Following \citet{gaschi-etal-2022}, we use a bilingual dictionary to extract matching pairs of words in translated sentences, discarding any ambiguity to obtain the most accurate pairs possible. 

It is worth noting that the accuracy of a nearest-neighbor search is not symmetric. We use the convention that an A-B alignment means that we look for the translation of each word of language A among its nearest neighbors. Two types of alignment can be evaluated: strong and weak alignment \cite{roy-etal-2020-lareqa}. Weak alignment is the expected way to compute alignment: when evaluating A-B weak alignment, we search a translation for a given word of A only among nearest-neighbors belonging to B. But with such an evaluation, there can be situations with highly measured alignment where representations from both languages are far apart with respect to intra-language similarity. Strong alignment remedies to this by including language A in the search space. With A-B strong alignment, we search a translation for a given word of A among its nearest-neighbors belonging to \emph{both} language B \emph{and} A. For a given pair of related words to be considered close enough, the word from language B must be closer to its translation in A than any other word from B \emph{and} A. We show in our experiments that strong alignment is more correlated with CTL than weak alignment.\footnote{Although strong alignment can be affected by synonyms, restricting the search space to the words from the sampled extracted pairs reduces the risk of founding a synonym.} 

\begin{figure*}[t]
     \subfloat[NLI, last layer]{
        \includegraphics[width=0.48\linewidth]{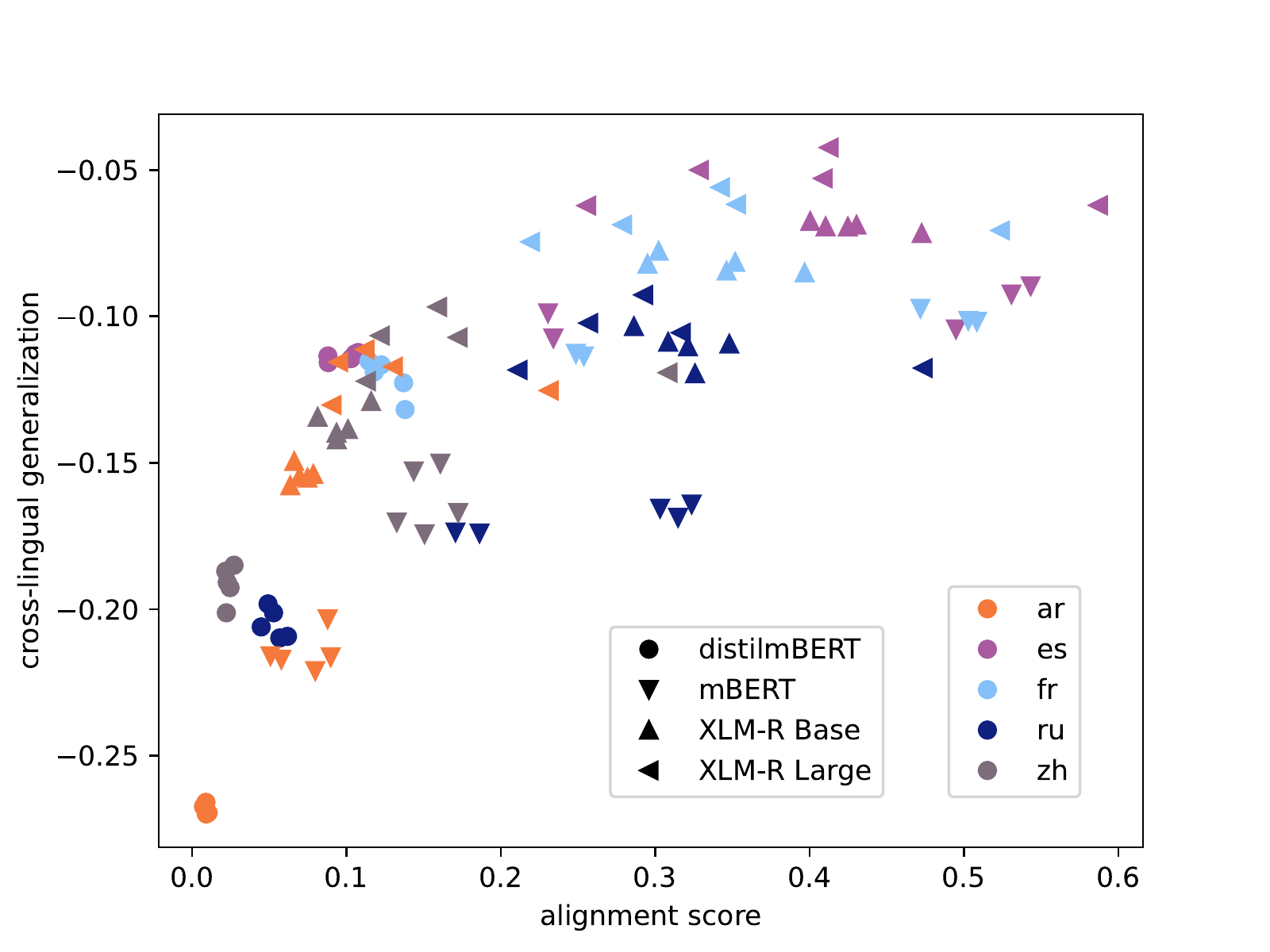}
         \label{subfig:corr_nli_last}
   }
    \subfloat[NLI, penultimate]{
    \includegraphics[width=0.48\linewidth]{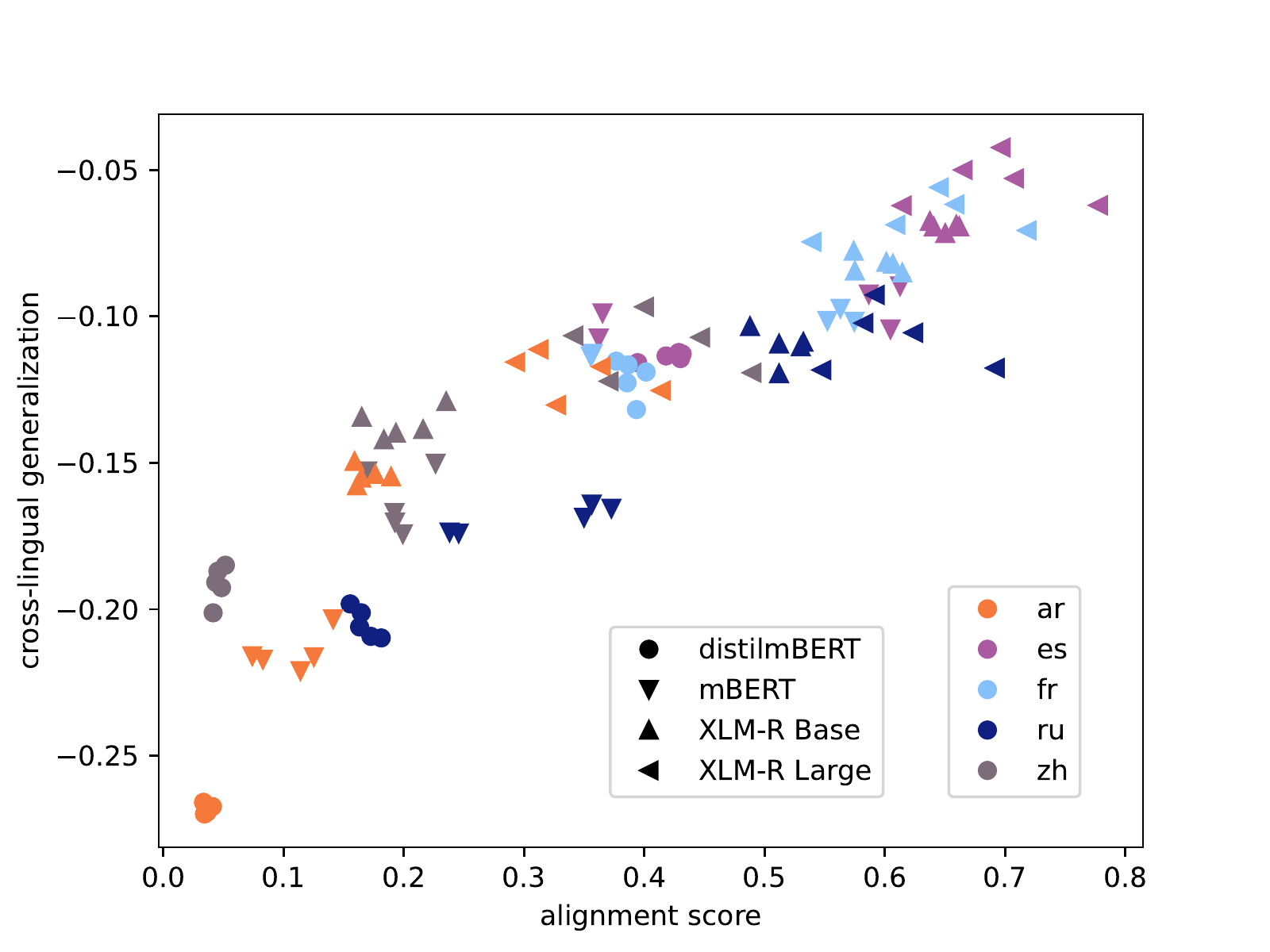}
        \label{subfig:corr_nli_penult}
    }
   
    \caption{Plot of CTL abilities against the English-target strong alignment measured for the last and penultimate layer after fine-tuning on NLI.}
    \label{fig:detailled_correlation} %TODO add best layer (layer with higher alignment)
\end{figure*}

\begin{figure}
    \centering
    \includegraphics[width=\linewidth]{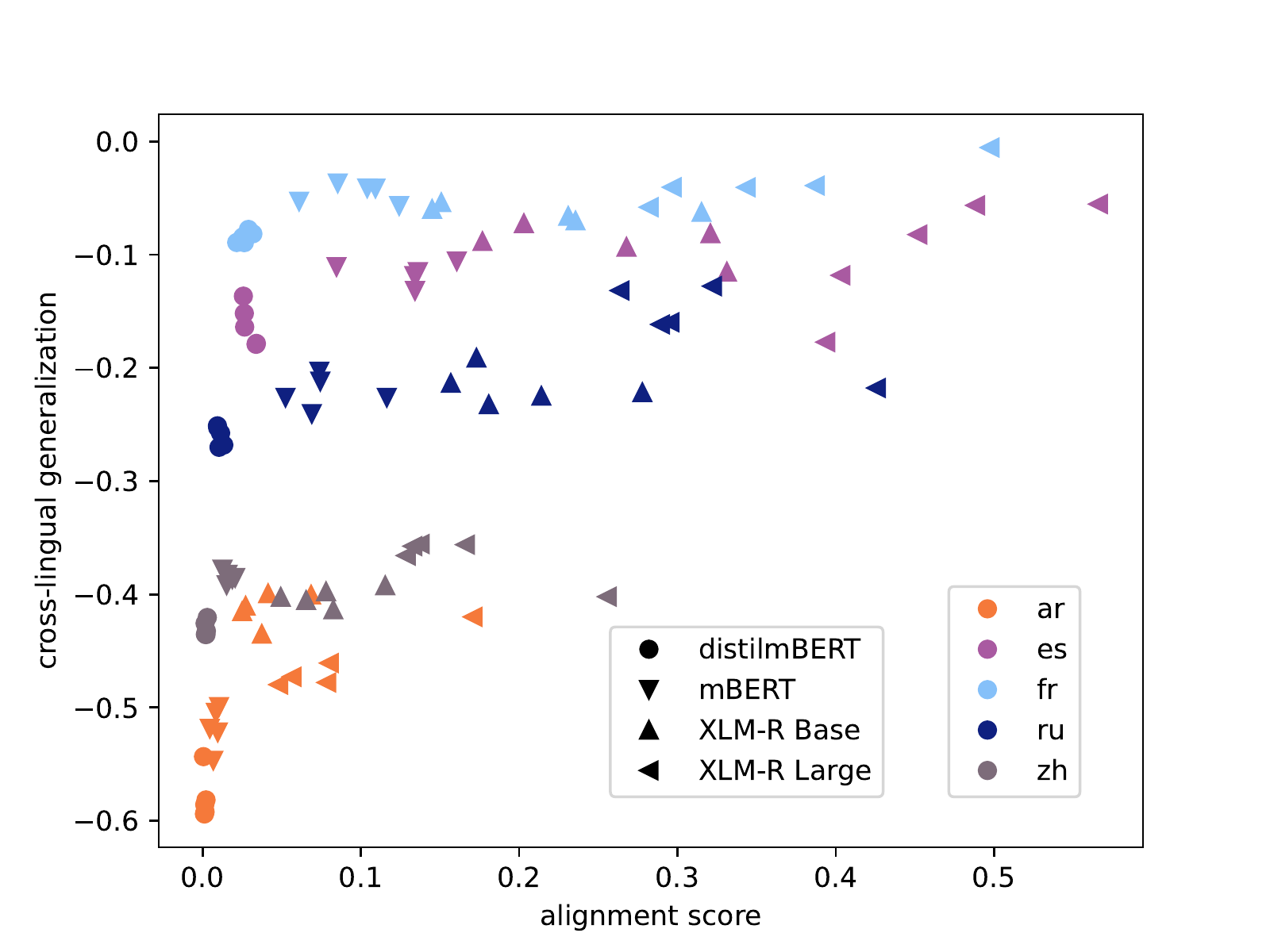}
    \caption{CLT abilities against English-target strong alignment for the last layer after fine-tuning on NER.}
    \label{subfig:corr_ner_last}
\end{figure}

\subsection{Realignment loss}

A realignment task consists in making the representations of related pairs closer to each other. The method used to extract related pairs for alignment evaluation can be used for computing the realignment loss. Following \citet{wu-dredze-2020-explicit}, we minimize a contrastive loss using the framework of \citet{chen-etal-2020-simple}, encouraging strong alignment for pairs within a batch. A batch is composed of a set of representations $\mathcal{H}$ of all words in a few pairs of translated sentences and a set $\mathcal{P}\subseteq \mathcal{H} \times \mathcal{H}$ containing the pairs of translated words extracted with a bilingual dictionary (or a word aligner). The realignment loss can then be written as:
\begin{equation}
    \begin{aligned}[b]
    \mathcal{L}_{\text{realign}} {=}  - \! \frac{1}{2|\mathcal{P}|} & \sum_{(s, t){\in} \mathcal{P}} \! \left[ \log \frac{
        \exp \! \left( \frac{\text{sim} (s, t)}{T} \right) 
    }{
        \sum_{\substack{h{\in}\mathcal{H} \\ h\neq s}} \exp \! \left( \frac{\text{sim} (s, h) }{T}\right)
    } \right. \\
    &\left. + \log \frac{
        \exp \! \left( \frac{\text{sim} (s, t)}{T} \right) 
    }{
        \sum_{\substack{h{\in}\mathcal{H} \\ h\neq t}} \exp \! \left( \frac{\text{sim} (h, t) }{T}\right)
    }
    \right]
    \end{aligned}
\end{equation}

For each pair $(s,t)$, the cosine similarity ($\text{sim}$) is compared to negative pairs (all other pairs in $\mathcal{H}$) using a softmax with a temperature hyper-parameter ($T=0.1$), following \citet{wu-dredze-2020-explicit}, bringing closer together translated pairs of words with respect to other pairs in the batch.

\subsection{Experimental details}\label{sec:exp_details}

We evaluate cross-lingual transfer with three multilingual tasks, the sizes of which are reported in Table \ref{tab:examples}:

\begin{itemize}
    \item Part-of-speech tagging (POS-tagging) with the Universal Dependencies dataset \cite{udpos}. Similarly to \citet{wu-dredze-2020-explicit}, we use the following treebanks: Arabic-PADT, English-EWT, Spanish-GSD, French-GSD, Russian-GSD, and Chinese-GSD.
    \item Named Entity Recognition (NER) with the WikiANN dataset \cite{pan-etal-2017-cross}.
    \item Natural Language Inference (NLI) with the XNLI dataset \cite{conneau-etal-2018-xnli}.
\end{itemize}

\begin{table}
    \centering
    \small
    \begin{tabular}{c|c|c|c}
         & POS & NER & XNLI \\
        \hline
        en-train & 12,543 & 20,000 & 392,703 \\
        en-dev & 2,002 & 10,000 & 2,490  \\
        en-test & 2,077 & 10,000 & 5,010 \\
        \hline
        ar-test & 680 & 10,000 & 5,010 \\
        es-test & 426 & 10,000 & 5,010  \\
        fr-test & 416 & 10,000 & 5,010 \\
        ru-test & 601 & 10,000 & 5,010  \\
        zh-test & 500 & 10,000 & 5,010\\
    \end{tabular}
    \caption{Number of examples}
    \label{tab:examples}
\end{table}

It must be noted that XNLI is the only dataset with translated test sets, and thus the only one for which the cross-lingual transfer metric is strictly comparable across languages. In our experiments, high correlation will nonetheless be observed between CTL and alignment for the two other tasks, suggesting that the CTL metrics is not so much affected by difference in size and domain between the test sets. 

Further details about implementation can be found in Appendix \ref{annex:exp_details} And in the source code\footnote{\url{https://github.com/posos-tech/multilingual-alignment-and-transfer}}.

\section{Correlation between alignment and CTL}\label{section:correlation}

\begin{table}[]
    \centering
    \small
    \begin{tabular}{l|l|cc|cc}
task & layer & \multicolumn{2}{c|}{weak} & \multicolumn{2}{c}{strong}\\
& & before & after & before & after \\
\hline
\multirow{2}{*}{POS} & last & 0.58 & 0.84 & 0.82 & 0.87 \\
 & penult & 0.78 & 0.84 & 0.87 & 0.86 \\
\hline
\multirow{2}{*}{NER} & last & 0.69 & 0.72 & 0.86 & 0.70 \\
 & penult & 0.82 & 0.71 & 0.87 & 0.82 \\
\hline
\multirow{2}{*}{NLI} & last & 0.51 & 0.75 & 0.86 & 0.82 \\
 & penult & 0.74 & 0.95 & 0.84 & 0.92 \\
\hline
    \end{tabular}
    \caption{Spearman's rank correlation of CTL with the English-target alignment produced by the last and penultimate layer before and after fine-tuning. Evaluation is done across 5 languages, 5 seeds and 4 models ($N=100$). All cells have p-value $<0.05$.}
    \label{tab:correlation}
\end{table}

% Hypothesis: alignment measured after fine-tuning is strongly linked to how the model will generalize to the target language, and should be comparatively more than alignment measured before

We measure the correlation between multilingual alignment and cross-lingual transfer (CTL) across models, languages and seeds. We also compare the correlation between alignment before fine-tuning and after fine-tuning with CTL and with different alignment measures.

% describe experiment for before/after correlation.
Spearman's rank correlation is measured between alignment before or after fine-tuning and CTL. The English-target alignment is computed for each target language with the method described in Section \ref{section:evaluation} and is compared with the transfer ability from English to that same target language with the metric described in Section \ref{section:evaluate_transfer}.

% Alignment is strongly correlated to cross-lingual transfer, whatever the task
Table \ref{tab:correlation} shows correlations between CTL and different types of alignment. It is computed separately for each different task (POS, NER, NLI), for the alignment at the last and second to last layer (last and penult), before and after fine-tuning on the given task, and with weak and strong alignment. Comparing other layers for models of different sizes is less relevant, since the correlation is computed across models with various number of layers. And a model-by-model analysis of the correlation with the alignment in various layers did not reveal contradictory results (cf. Appendix \ref{section:correlation_breakdown}). Each correlation value is obtained from 100 samples with four different models (distilmBERT, mBERT, XLM-R Base and Large), five target languages (Arabic, Spanish, French, Russian and Chinese) and five seeds for initialization of the classification head and shuffling of the fine-tuning data.

Results show that strong alignment is better correlated to cross-lingual transfer than weak alignment. With the exception of two tasks after fine-tuning (NER and NLI), strong alignment has a marginally higher correlation with CTL. This is particularly noticeable when looking at alignment before fine-tuning on the last layer, going from a correlation between 0.51 and 0.69 for weak alignment to one ranging from 0.82 to 0.86 for strong alignment.

% NLI antepenultimate
Tab. \ref{tab:correlation} also shows that for NLI, the alignment on the penultimate layer seems better correlated to cross-lingual transfer than with the last layer. A relatively important gap in correlation is measured between the last and the second-before-last layer for all cases except for strong alignment before fine-tuning. The fact that alignment on the penultimate layer would correlate better than the last  for NLI can be explained by the sentence-level nature of the task. For sentence classification tasks, the classification head reads only the representation of the first token of the last layer, which is computed from the representations of all the tokens at the previous layer, leading to a pooling of the penultimate layer.

% Hard to conclude about before/after and forward/backward

Despite the different values observed, there seems to be no significant difference between correlation for alignment measured before and after fine-tuning, and a careful analysis of confidence interval obtained with bootstrapping  \cite{Efron1994AnIT} can confirm this (cf. Appendix \ref{section:ci_correlation} for detailed results). % TODO compute correlation between before and after

%TODO see if we remove the rest from there to the end of the section

Fig. \ref{fig:detailled_correlation} shows the relation between CTL and English-target strong alignment measured after fine-tuning measured in four situations to further illustrate the link between alignment and transfer. 

Fig. \ref{subfig:corr_nli_penult} shows one of the cases with higher correlation (0.92). The correlation seems to hold well across models (forms) and languages (colors). However, for a given model and language, the random seed for fine-tuning seems to be detrimental to the correlation, although at a small scale. Hence, alignment might not be the only factor to affect cross-lingual generalization as the model initialization or the data shuffling seems to play a smaller role. 

Fig. \ref{subfig:corr_ner_last} shows a case with one of the lowest correlations between strong alignment and CTL (0.70). It seems that models and initialization seeds have a higher impact on alignment than on CTL. For example, in the case of English-French alignment (green), CTL is between 0.0 and -0.1, whatever the model and seed, not overlapping with other target-English language pairs, but alignment varies between approximately 0.05 and 0.5, overlapping with all other language pairs. Interestingly, the penultimate layer has a higher accuracy (0.82), suggesting that for NER the last layer is not necessarily the one for which alignment correlates the most with CTL. 

For two of the three tested tasks (NER and POS-tagging), it must be noted that the CTL metric is not strictly comparable across languages since the test sets for each language are of different domains and sizes (cf. Section \ref{sec:exp_details}). However, for the third task (NLI), each test set is a translation of the English one, and thus the CTL metric is strictly comparable in that case. This might explain why correlations are higher for the NLI task than the others. Nevertheless, the observed correlation for the two other tasks is still significantly high, which suggests that the general tendency might not be affected by the differences in domains and sizes in the test sets.

% Transition: correlation is not causality, so can we build realignment methods that shows that improving aligment is indeed beneficial to CTL ?

\section{The impact of fine-tuning on alignment}\label{section:fine-tuning}

\begin{figure*}[t]
    \centering
    \subfloat[distilmBERT POS]{
    	\includegraphics[width=0.48\textwidth]{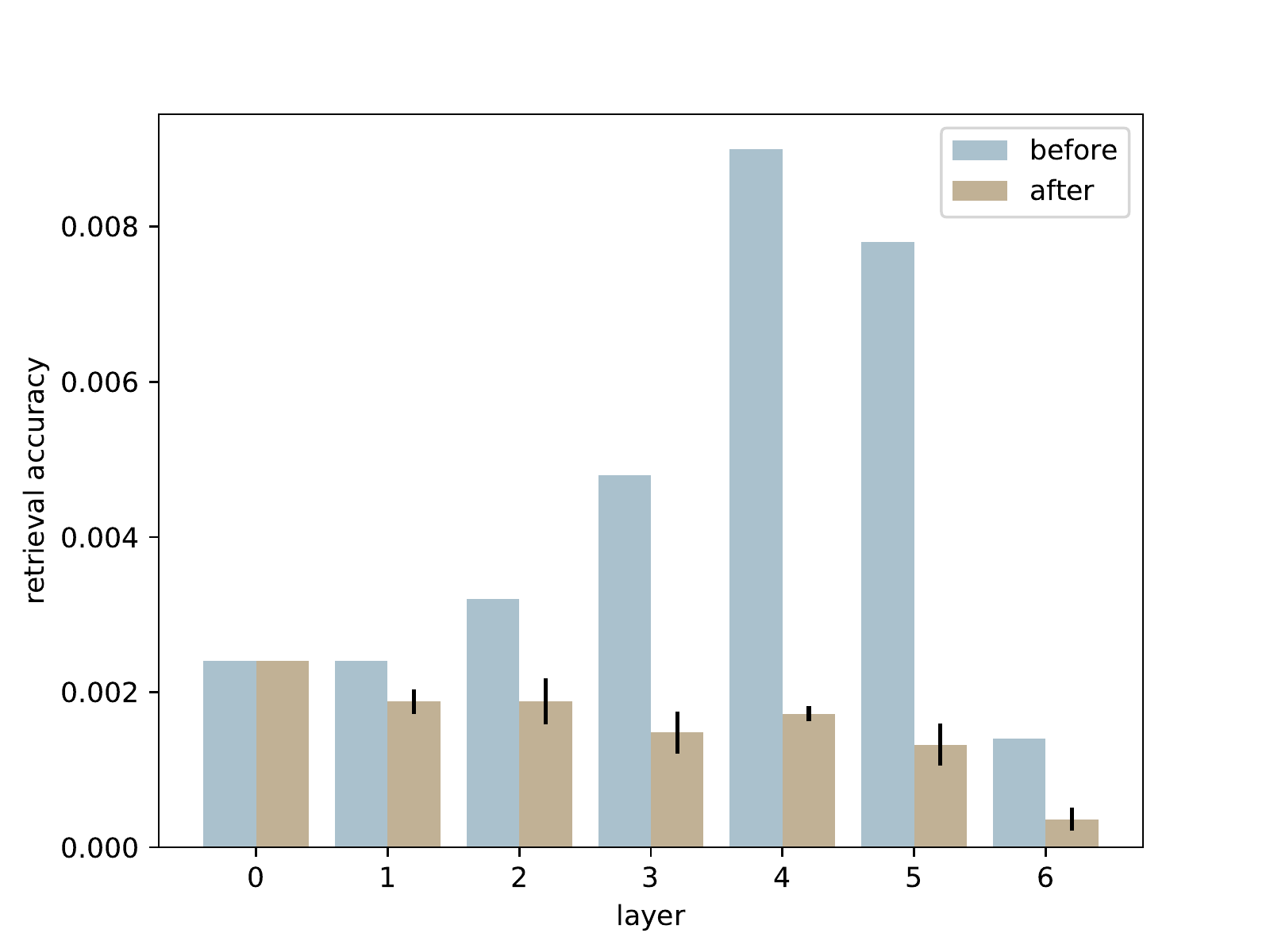}
        \label{fig:pos_distil}
    }
    \subfloat[XLM-R Large POS]{
    	\includegraphics[width=0.48\textwidth]{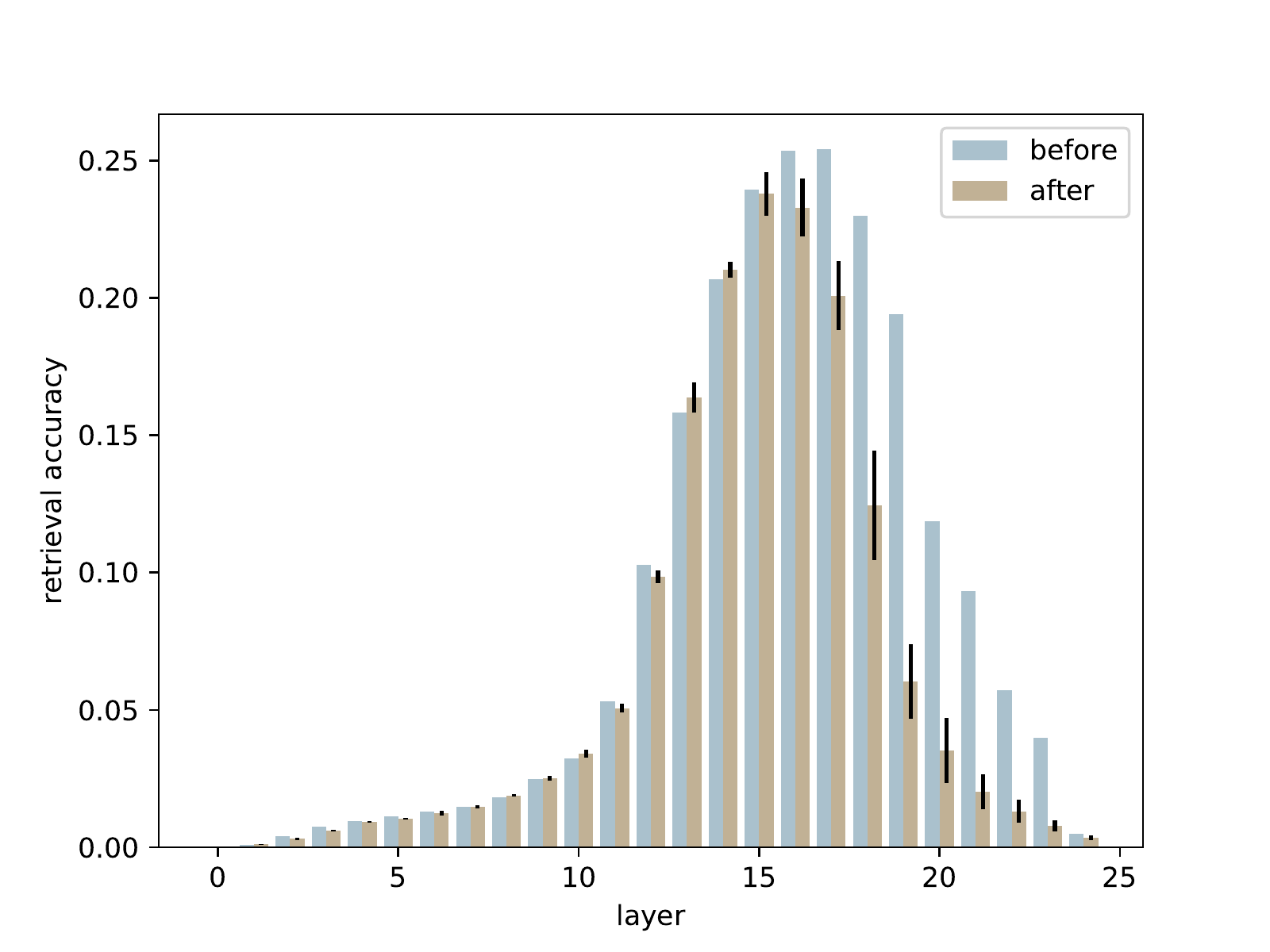}
        \label{fig:pos_large}
    }
    \caption{Evolution across layers of English-Arabic alignment before and after fine-tuning of distilmBERT and XLM-R Large on POS-tagging, starting at 0 for the embedding layer.} %TODO put them with same scale ?
    \label{fig:drop_layer}
\end{figure*}
\begin{figure}
		\centering
        \includegraphics[width=0.48\textwidth]{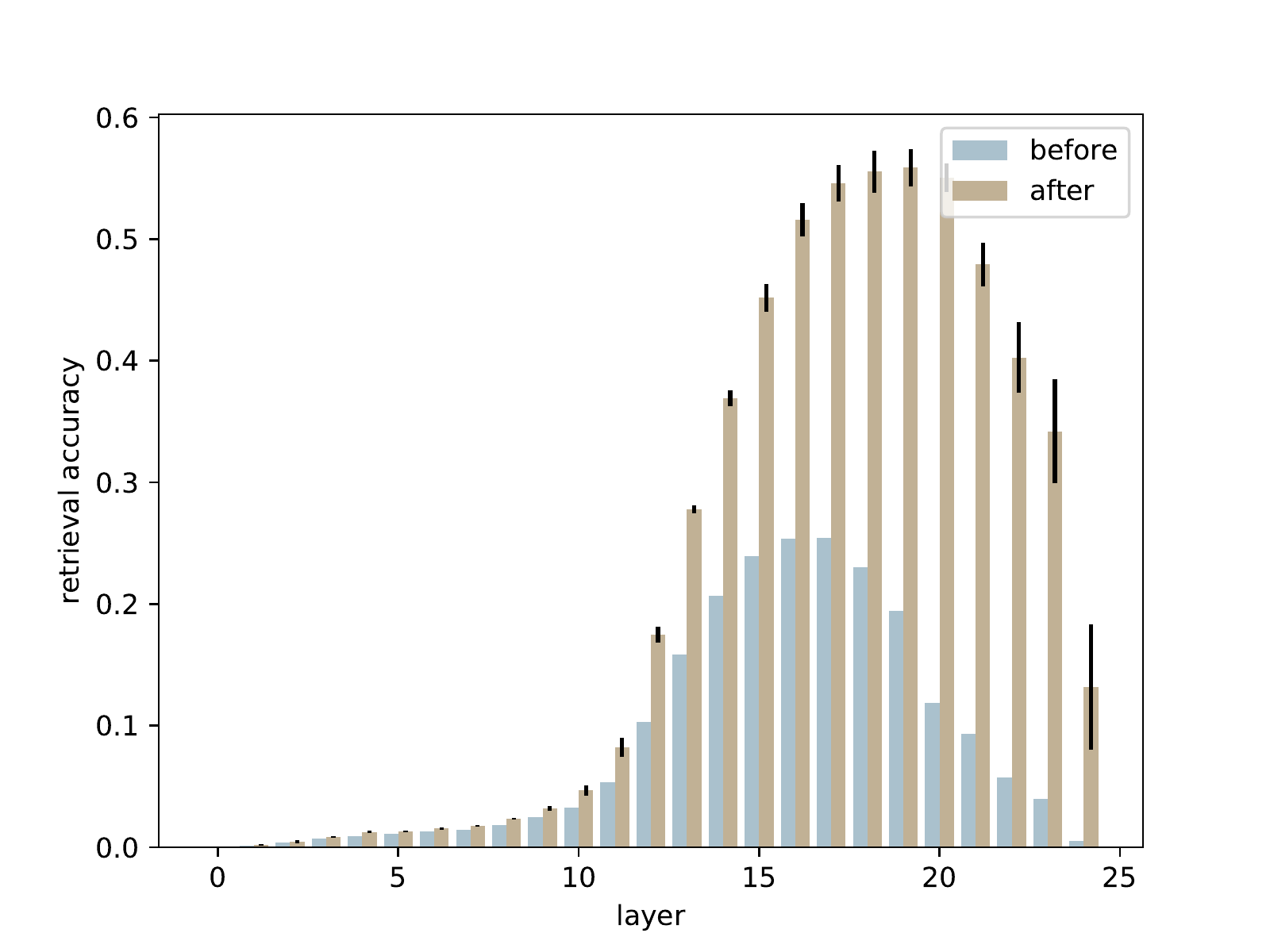}
     
        \caption{Alignment across layers of English-Arabic before and after fine-tuning of XLM-R Large on NLI.}
           \label{fig:nli_large}
\end{figure}

% What we want to do: check whether representation built by MLLMs are aligned
To study the link between alignment and cross-lingual transfer (CTL), we also look at the impact of fine-tuning over alignment. We've already shown that strong alignment is highly correlated with CTL. However, we weren't able to conclude whether alignment measured before or after fine-tuning was better correlated to CTL abilities. To understand the difference between both measures, we study in this section the impact of fine-tuning on the alignment of MLLMs representations. We use the same fine-tuning runs as in the previous section (\ref{section:correlation}).

% Analyze the dramatic drop in accuracy Tab. 1
Tab. \ref{tab:drop_alignment} shows the relative variation in alignment before and after fine-tuning for all tasks and models tested and for three languages for clarity (complete results in Appendix \ref{section:detailed_drop}). The relative difference is built in the same way as the cross-lingual transfer evaluation (Eq. \ref{eq:transfer}). Negative values indicate a drop in alignment. Alignment is measured at the last layer. Fig. \ref{fig:drop_layer} and \ref{fig:nli_large} show a breakdown by layer for a few cases.
\looseness=-1

For certain combinations of models and tasks, fine-tuning is detrimental to multilingual alignment. distilmBERT and mBERT mainly show a decrease in alignment for POS-tagging and NER, and smaller improvements than other models on NLI. However, POS-tagging is the only of the three tasks which shows dramatic drops where alignment can be reduced by as much as 96\%.

The drop in alignment can be explained by catastrophic forgetting. If the model is only trained on a monolingual task, it might not retain information about other languages or about the link between English and other languages. 

\begin{table}[]
    \centering
    \small
    \begin{tabular}{l|l|c|c|c}
task & model & en-ar & en-es & en-ru\\                                                                                                                            
\hline\multirow{4}{*}{POS}& distilmBERT &  -0.74 &  -0.86 &   -0.87\\                                                                                             
& mBERT &  -0.90 &  -0.86 &    -0.95\\                                                                                                                            
& XLM-R Base &  -0.43 &  -0.46 &   -0.70\\                                                                                                                        
& XLM-R Large &  -0.30 &  0.23 &  -0.44\\                                                                                                                         
\hline\multirow{4}{*}{NER}& distilmBERT &  0.00 &  -0.61 &   -0.33\\                                                                                              
& mBERT &  -0.28 &  -0.36 &    -0.27\\                                                                                                                           
& XLM-R Base &  5.88 &  0.22 &   1.32\\
& XLM-R Large &  16.34 &  2.22 &   3.10\\
\hline\multirow{4}{*}{NLI}& distilmBERT &  5.49 &  0.30 &  2.28\\                                                                                                                              
& mBERT &  5.65 &  0.99 &   1.45\\                                                                                                                                                              
& XLM-R Base &  11.17 &  1.01 &   2.67\\                                                                                                                                                       
& XLM-R Large &  25.36 &  1.78 &   2.99\\                           
\hline

    \end{tabular}
    \caption{Relative variation of strong alignment at the last layer before and after fine-tuning for different fine-tuning tasks (nearest-neighbor search accuracy after fine-tuning minus accuracy before).}
    \label{tab:drop_alignment} %TODO add complete results in Appendix
    %TODO see if we might not need strong alignment instead (since it is more highly correlated) maybe invert the two parts 
\end{table}

What is more surprising is the increase in alignment obtained in other cases. XLM-R Base and Large, which are larger models than mBERT and distilmBERT, have a relative increase that can go as high as 25.36 on the NLI task for distant languages. And although these increases are from a small alignment measure, we still observe a large increase for middle layers where the initial alignment is already quite high (cf. Fig. \ref{fig:nli_large}). 

The alignment of larger models being less harmed by fine-tuning is coherent with the fact that those same larger models have been shown to have better CTL abilities. Fig. \ref{fig:drop_layer} shows that more layers seem to mitigate the potentially negative impact of fine-tuning on alignment, as it affects mainly the layers closest to the last one and as the initial alignment measure is globally higher for XLM-R than distilmBERT (before fine-tuning: $\approx$0.25 against $\approx$0.008).
\looseness=-1
%TODO add ref to deeper models = better multilingual model

Giving a definitive answer as to why different tasks have different impacts on alignment might need further research. But one could already argue that each task corresponds to different levels of abstraction in NLP. Tasks with a low level of abstraction like POS-tagging might rely on the word form itself and thus on more language-specific components of the representations, which when enhanced, decreases alignment. On the other hand, NLI has a higher level of abstraction, requiring the meaning rather than the word form, which might be encoded in deeper layers \cite{tenney-etal-2019-bert} which are more aligned.

% Conclude on drop in alignment and ask questions for the following
% - does a descreased alignment means that the cross-lingual transfer abilities will be harmed ?
% - How can it inform realignment methods and choice of model architecture ?
Fine-tuning MLLMs on a downstream task has an impact on the multilingual alignment of the representations produced by the model. For "smaller" language models, it is systematically detrimental, as well as for certain tasks like POS-tagging. This might explain why some realignment methods might not work for all models nor all tasks \cite{wu-dredze-2020-explicit}.

\section{Impact of realignment on cross-lingual transfer}\label{section:realignment}

We have already shown that the correlation between multilingual alignment and cross-lingual transfer (CTL) is high (Section \ref{section:correlation}). But we do not know whether they are more directly linked. In this section, we try to identify the conditions under which improving alignment in multilingual models leads to improvement in CTL. 

Sequential realignment is the usual way to perform realignment: realignment steps are performed on the pre-trained model before fine-tuning. We propose to compare it with joint alignment, where we optimize simultaneously for the realignment and the downstream task (more details in Appendix \ref{section:joint_realignment}), to try and identify whether alignment before or after fine-tuning is more strongly related to CTL.

In the same settings as the previous experiments (tasks, models and languages, and number of seeds), we fine-tune models in English with different realignment methods and evaluate CTL on different languages. Following a similar setting as \cite{wu-dredze-2020-explicit}, realignment data from the five pairs of languages (English-target) is interleaved to form a single multilingual realignment dataset. Models are fine-tuned on POS-tagging or NER for five epochs and 2 epochs for NLI because its training data is larger. We use the opus100 translation dataset \cite{zhang-etal-2020-improving} from which we extract pairs of words using bilingual dictionaries. We also tested with the multiUN translation data \cite{ziemski-etal-2016-united}, which conditioned our choice of languages, and with other ways to extract alignment pairs: FastAlign \cite{dyer-etal-2013-simple} and AWESOME-align \cite{dou-neubig-2021-word}. Changing the translation dataset does not fundamentally change the results, and using probabilistic alignment tools made realignment methods less effective. The results presented in this section were handpicked for the sake of clarity, but the reader can refer to Appendix \ref{section:detailed_realignment}. 

\begin{table}
    \centering
    \small
    \begin{tabular}{l|c|c|c}
        & POS & NER & NLI\\
        \hline  
        distilmBERT & \textit{ 73.1 } & \textit{ 57.4 } & \textit{ 64.2 } \\
        + before & \textbf{ 78.0 } & 58.9 & \cellcolor{gray!30} 64.3 \\
        + joint & 77.9 & \textbf{ 59.6 } & \cellcolor{gray!30} \textbf{ 65.0 } \\
        \hline
        mBERT & \textit{ 74.3 } & \textit{ 62.2 } & \textit{ 69.7 } \\
        + before & 78.2 & \cellcolor{gray!30} 62.7 & \cellcolor{gray!80} 68.7 \\
        + joint & \textbf{ 78.3 } & \textbf{ 64.8 } & \cellcolor{gray!30} \textbf{ 70.0 } \\
        \hline
        XLM-R base & \textit{ 78.8 } & \textit{ 60.4 } & \textit{ 74.0 } \\
        + before & \textbf{ 79.9 } & \textbf{ 63.9 } & \cellcolor{gray!30} 72.7 \\
        + joint & 79.4 & 63.0 & \cellcolor{gray!30} \textbf{ 74.6 } \\
        \hline
        XLM-R large & \textit{ 79.6 } & \textit{ 65.0 } & \textit{ 80.0 } \\ 
        \hline
    \end{tabular}
    \caption{Condensed results of the controlled experiment comparing joint and sequential realignment using a bilingual dictionary. Light gray indicates a difference with baseline lower than its standard deviation. Dark gray indicates lower than baseline minus standard deviation.
    \looseness=-1}
    \label{tab:controlled}
\end{table}
\begin{table}
    \centering
    \small
    \begin{tabular}{l|c|c|c|c|c}
       POS & ar& es & fr & ru & zh\\
    \hline
    distilmBERT & \textit{51.0} & \textit{84.1} & \textit{85.3} & \textit{81.2} & \textit{64.1} \\
    + before  & 65.5 & \textbf{86.5} &\textbf{85.8} &  \textbf{84.7} & \textbf{67.4} \\
    + joint  & \textbf{66.8} & \textbf{85.8} & 86.5 & 84.1 & 66.4 \\
    \hline                                                                                                                                                                                                                                                                          
XLM-R base & \textit{62.5} &  \textit{86.6} & \textit{86.9} & \textit{\textbf{86.9}} & \textit{70.9} \\                                                                                  
+ before  & \textbf{67.3}& \cellcolor{gray!30}\textbf{86.9} &  \textbf{87.3} & \cellcolor{gray!30}86.8 & \cellcolor{gray!30}\textbf{71.2} \\                                        
+ joint & 66.6 & \cellcolor{gray!30}86.6 & 87.2 &  \cellcolor{gray!80}86.0 & \cellcolor{gray!30}70.6  \\
\hline
XLM-R large & \textit{\textbf{65.1}}& \textit{\textbf{87.0}}& \textit{\textbf{87.5}} & \textit{\textbf{87.0}} & \textit{\textbf{71.5}}\\
\hline

    \end{tabular}
    \caption{Breakdown of realignment results for some languages and distilmBERT and XLM-R.}\label{tab:controlled_lang_and_models}
\end{table}    

Condensed results are reported on Tab. \ref{tab:controlled}, averaged on the five languages. A breakdown by languages for the POS-tagging task and two models is shown on Tab. \ref{tab:controlled_lang_and_models}. It shows that realignment methods improve performance only on certain tasks, models and language pairs.

Realignment methods, either sequential or joint, provide significant improvement for all models for the POS-tagging task, but less significant ones for NER, and no significant improvement for NLI. The positive impact of realignment on cross-lingual transfer seems to be mirrored by the negative impact of fine-tuning over alignment. Indeed, POS-tagging is also the task for which fine-tuning is the most detrimental to multilingual alignment, as shown in the previous section. 

The same parallel can be drawn for models. distilmBERT is the model that benefits the most from realignment. It is also the one whose alignment suffers the most from fine-tuning. Smaller multilingual models seem to benefit more from realignment, as well as they see their multilingual alignment reduced after fine-tuning. In the same way that fine-tuning mainly affects the deeper layers, it is possible that realignment might affect only those deeper layers. This would mean that most layers would have their alignment significantly improved for small models like distilmBERT (6 layers), while larger models might be only superficially realigned.

Finally, besides tasks and models, it can also be observed that the impact of realignment varies across language pairs (Tab. \ref{tab:controlled_lang_and_models}). Although we did not test on many language pairs, results are coherent with the idea that realignment methods tend to work better on distant pairs of languages \cite{kulshreshtha-etal-2020-cross}. 

On a side note, our controlled experiment does not allow us to conclude whether it is more important to improve alignment before fine-tuning or after. It seems that alignment measured before and after fine-tuning are equally important to cross-lingual transfer.
 
Realignment methods unsurprisingly provide better results when the alignment is lower, be it before or after fine-tuning. Distant languages and small models have lower alignment, and POS-tagging is a task where alignment decreases after fine-tuning. Realignment helps only up to a certain point where representations are already well aligned, and CTL gives already good results. For distilmBERT on POS-tagging for transfer from English to Arabic, it provides a +15.8 improvement over baseline, even outperforming XLM-R Large by 1.7 points. In such conditions, realignment is an interesting alternative to scaling for multilingual models.
\looseness=-1

If realignment succeeds in some favorable conditions, then how can we explain that realignment methods were shown to not be significantly improving CTL on several tasks, including POS-tagging \cite{wu-dredze-2020-explicit}? Firstly, to the best of our knowledge, realignment was never tried on distilmBERT or other models of equivalent size. Secondly, Tab. \ref{tab:controlled_avg_pairs} shows that it might be partly due to an element of the realignment methods that was overlooked: the source of related pairs of words.

% - pair extraction
\begin{table}
    \centering
    \begin{tabular}{l|c|c}
        & POS & NER\\
\hline
XLM-R base & \textit{ 78.8 } & \textit{ 60.4 } \\
+ before fastalign & \cellcolor{gray!30} 78.6 & \cellcolor{gray!30} 61.4 \\
+ before awesome & \cellcolor{gray!30} 78.6 & \cellcolor{gray!30} 62.0 \\
+ before dico & \textbf{ 79.9 } & \textbf{ 63.9 } \\
+ joint fastalign & \cellcolor{gray!80} 78.0 & \cellcolor{gray!30} 62.1 \\
+ joint awesome & \cellcolor{gray!80} 77.8 & \cellcolor{gray!30} 62.3 \\
+ joint dico & 79.4 & 63.0 \\
\hline

    \end{tabular}
    \caption{Average CTL abilities for XLM-R with different type of realignment.}
    \label{tab:controlled_avg_pairs}
\end{table}

The way pairs are extracted seems to be crucial to the success of realignment methods. Tab. \ref{tab:controlled_avg_pairs} shows the effect of different types of pairs extraction in realignment methods. Realignment methods using pairs extracted with FastAlign or AWESOME-align do not provide significant improvements over the baseline, whereas using a bilingual dictionary does. Using a bilingual dictionary might be more accurate for extracting translated pairs \cite{gaschi-etal-2022}. Another explanation could be that the type of words contained in a dictionary might help since it might contain more lexical words holding meaning and fewer grammatical words.

\section{Conclusion}

We have shown that multilingual alignment, measured using a nearest-neighbor search among translated pairs of contextualized words, is highly correlated with the cross-lingual transfer abilities of multilingual models (or at least multilingual Transformers). Strong alignment was also revealed to be better correlated to cross-lingual transfer than weak alignment.

Then we investigated the impact of fine-tuning (necessary for cross-lingual transfer) on alignment as well as the impact of realignment methods on cross-lingual transfer. Fine-tuning was revealed to have a very different impact on alignment depending on the downstream task and the model. Where lower-level tasks seemed to have the most impact and smaller models seemed to be the most affected. Conversely, realignment methods were shown to work better on those same tasks and models. Ultimately, realignment works unsurprisingly better when the baseline alignment (before or after fine-tuning) is lower. 

We also showed that using a bilingual dictionary for extracting pairs for realignment methods improves over the commonly used FastAlign and over a more precise neural aligner (AWESOME-align).

It's worth noting that realignment works particularly well for a small model like distilmBERT (66M parameters), allowing it in some cases to obtain competitive results with XLM-R Large (354M parameters). This advocates for further research on realignment for small Transformers to build more compute-efficient multilingual models.

Finally, further research is needed to investigate additional questions, like whether cross-lingual transfer is more directly linked to alignment before or after fine-tuning, or to alignment at certain layers for certain tasks. To answer these questions, more large-scale experiments could be performed on more tasks and especially on more languages to obtain correlation values with smaller confidence intervals. 

\section{Limitations}

We worked with only five language pairs, all involving English and another
language: Arabic, Spanish, French, Russian and Chinese. This is due to using the
multiUN dataset \cite{ziemski-etal-2016-united} for evaluating alignment and
performing realignment. We also used the opus100 dataset
\cite{zhang-etal-2020-improving}, which contains more pairs and is the dataset
that eventually figured in our paper, but we stuck to the same language pairs
for a fair comparison with multiUN in Appendix
\ref{section:detailed_realignment}. This narrow choice of language limits our
ability to understand why realignment methods work well for some languages and
not others. And we believe that making a similar analysis with more language
pairs, not necessarily involving English, might answer more questions. 

We chose a strong alignment objective with contrastive learning for our
realignment task. Several other objectives could have been tried, like learning
an orthogonal mapping between representations \cite{wang-etal-2019-cross} or
simply using a $\ell_2$-loss to collapse representations together
\cite{Cao2020Multilingual}, but both methods require an extra regularization
step \cite{wu-dredze-2020-explicit} since they do not rely on any negative
samples. For the sake of simplicity, we focused on a contrastive loss, as trying
different methods would have led to an explosion in the number of runs for the
controlled experiment. This also explains why we used the same hyper-parameters
and pre-processing steps of \citet{wu-dredze-2020-explicit}. A more thorough
search for the optimal parameters might lead to better
results.

\section{Acknowledgements}

We would like to thank the anonymous reviewers for their comments, as well as Shijie Wu, who kindly explained some details in the implementation of his paper \citet{wu-dredze-2020-explicit}. We are also grateful for discussion and proof-reading brought by our colleagues at Posos: François Plesse, Xavier Fontaine and Baptiste Charnier.

Experiments presented in this paper were carried out using the Grid'5000 testbed, supported by a scientific interest group hosted by Inria and including CNRS, RENATER and several Universities as well as other organizations\footnote{see https://www.grid5000.fr}. 

\bibliography{anthology,custom}
\bibliographystyle{acl_natbib}

\newpage
\clearpage

\appendix

\section{Joint realignment} \label{section:joint_realignment}

Existing realignment methods proceed in a sequential manner. The pre-trained model is first optimized for the realignment loss, before any fine-tuning. This assumes that the alignment before fine-tuning is positively linked to the cross-lingual transfer abilities of the model and that improving alignment before fine-tuning will improve transfer. However, fine-tuning itself might have an impact on alignment \cite{eftimov-etal-2022-impact}. 

To compare the importance of alignment before and after fine-tuning for CTL, we introduce a new realignment method where realignment and fine-tuning are performed jointly. We optimize simultaneously for a realignment loss and the fine-tuning loss. In practice, for each optimization step, we compute the loss $\mathcal{L}_{\text{task}}$ for a batch of the fine-tuning task and the loss $\mathcal{L}_{\text{realign}}$ for a batch of the alignment data. The total loss for each backward pass is then written as:
\begin{equation}
    \mathcal{L} = \mathcal{L}_{\text{task}} + \mathcal{L}_{\text{realign}}
\end{equation}
This joint realignment can be framed as multi-task learning. The fine-tuning task would be the main task and the realignment task an auxiliary one. There are more elaborate methods for training a model with an auxiliary task \cite{liebel-korner-2018,yunshu-etal-2018,zhang-etal-2018-fine,Liu-etal-2019-self}
but our aim is to propose the simplest method possible to compare joint and sequential alignment in a controlled setting.

\section{Experimental details}\label{annex:exp_details}

\subsection{Scientific artifacts used}

We relied on the following scientific Python packages for our experiments: the HuggingFace's libraries \texttt{transformers} \cite{wolf-etal-2020-transformers}, \texttt{datasets} \cite{lhoest-etal-2021-datasets} and \texttt{evaluate}\footnote{\url{https://huggingface.co/docs/evaluate/index}}, \texttt{PyTorch} \cite{torch_paper}, \texttt{NLTK} \cite{bird2009natural} and its implementation of the Stanford Chinese Segmenter \cite{tseng-etal-2005-conditional}, \texttt{seqeval} \cite{seqeval} for evaluating NER, \texttt{NumPy} \cite{harris2020array}, and AWESOME-align \cite{dou-neubig-2021-word}, FastAlign \cite{dyer-etal-2013-simple}, and MUSE dictionaries \cite{conneau2017word} for extracting alignment pairs.

We used the following datasets: two translation datasets for building evaluating alignment and realigning, multiUN \cite{ziemski-etal-2016-united} and opus100 \cite{zhang-etal-2020-improving}, XNLI \cite{conneau-etal-2018-xnli}, the Universal Dependencies dataset \cite{udpos} for POS-tagging, and the WikiANN dataset for NER \cite{pan-etal-2017-cross}.

Finally, we worked with four different models: distilmBERT, which was released with distilBERT \cite{Sanh2019DistilBERTAD}, mBERT, which was released with BERT \cite{devlin-etal-2019-bert} and XLM-R Base and Large \cite{conneau-etal-2020-unsupervised}.
\begin{table}[]
    \centering
    \begin{tabular}{c|c}
        model & \# parameters  \\
        \hline
        distilmBERT & 66M\\
        mBERT & 110M \\
        XLM-R Base & 125M \\
        XLM-R Large & 345M \\
    \end{tabular}
    \caption{Number of parameters}
    \label{tab:params}
\end{table}

\subsection{Multilingual alignment data}

From a translation dataset, pairs were extracted either using a bilingual dictionary, following \cite{gaschi-etal-2022}, with FastAlign \cite{dyer-etal-2013-simple} or AWESOME-align \cite{dou-neubig-2021-word}. For FastAlign, we produced alignments in both direction and symmetrize with the \texttt{grow-diag-final-and} heuristic provided with FastAlign, following the setting of \citet{wu-dredze-2020-explicit}. For all methods of extraction, we kept only one-to-one alignment and discard trivial cases where both words are identical, again following \citet{wu-dredze-2020-explicit}.

\subsection{Experimental setup}

We performed two experiments: 

\begin{enumerate}
    \item Fine-tuning all models on all tasks for 5 epochs and measuring alignment before and after fine-tuning. This experiment provided the results for Section \ref{section:correlation} and \ref{section:fine-tuning}.
    \item Performing different realignment methods before fine-tuning for 5 epochs (or 2 for XNLI), providing results for section \ref{section:realignment}.
\end{enumerate}

For both experiments, we reused the experimental setup from \citet{wu-dredze-2020-explicit}. Fine-tuning on a downstream task is done with Adam, with a learning rate of \texttt{2e-5} with a linear decay and warmup for 10\% of the steps. Fine-tuning is performed on 5 epochs, and 32 batch size, except for XNLI in the second experiment, where we trained for 2 epochs, which still leads to more fine-tuning steps than any of the two other tasks (cf. Table \ref{tab:examples}).

For the realignment methods, still following \citet{wu-dredze-2020-explicit}, we train in a multilingual fashion, where each batch contains examples from all target languages. However, we use the same learning rate and schedule as for fine-tuning for a fair comparison between joint and sequential realignment, since the same optimizer is used for fine-tuning and realignment when performing joint realignment. We use a maximum length of 96 like \citet{wu-dredze-2020-explicit} but a batch size of 16 instead of 128 because of limited computing resources. 

\subsection{Discussion on the number of realignment samples}

It is worth noting that our method uses fewer realignment samples. Since we alternate batches of 16 realignment samples and batches of 32 fine-tuning samples for joint realignment, this fixes the number of realignment samples we will use for a specific downstream task, for a fair comparison. This gives 31,358 sentence pairs for POS-tagging, 50,000 for NER, and 392,703 for NLI. For comparison, \citet{wu-dredze-2020-explicit} used 100k steps of batches of size 128. The number of realignment samples 
used could have been a factor explaining why realignment works well for POS-tagging and less for NER and NLI, and why \citet{wu-dredze-2020-explicit} do not find that realignment methods improve results significantly on any task. It could be argued that training on too many realignment samples might hurt performances. However, when testing on the POS-tagging task, we found that the number of realignment samples did not have significant impact on performances.

\subsection{Computational budget}

The first experiment was performed on Nvidia A40 GPUs for an equivalent of 3 days for a single GPU (including all models, tasks and seeds). For the second experiment, training (fine-tuning and/or realignment) was performed on various smaller GPUs (RTX 2080 Ti, GTX 1080 Ti, Tesla T4) for distilmBERT, mBERT and XLM-R Base, and on a Nvidia A40 for XLM-R Large. The experiment took more than 10 GPU-days on the smaller GPUs, combining all models, realignment methods (including baseline), random seeds, translation datasets and pairs extraction methods. For XLM-R Large, for which we only trained the baseline, it still required 30 GPU-hours on Nvidia A40. 

\begin{table*}[t]
    \centering
    \begin{tabular}{l|l|cc|cc}
task & layer & \multicolumn{2}{c|}{en-X} & \multicolumn{2}{c}{X-en}\\
& & before & after & before & after \\
\hline
\multirow{2}{*}{POS} & last & 0.58 (0.43 - 0.70) & 0.84 (0.77 - 0.89) & 0.63 (0.48 - 0.74) & 0.83 (0.74 - 0.89) \\
 & penult. & 0.78 (0.68 - 0.85) & 0.84 (0.76 - 0.89) & 0.80 (0.71 - 0.87) & 0.85 (0.79 - 0.90) \\
 \hline
\multirow{2}{*}{NER} & last & 0.69 (0.55 - 0.79) & 0.72 (0.59 - 0.81) & 0.75 (0.64 - 0.83) & 0.84 (0.73 - 0.89) \\
 & penult. & 0.82 (0.73 - 0.88) & 0.71 (0.58 - 0.81) & 0.88 (0.83 - 0.92) & 0.72 (0.58 - 0.82) \\
 \hline
\multirow{2}{*}{NLI} & last & 0.51 (0.32 - 0.67) & 0.75 (0.61 - 0.85) & 0.54 (0.36 - 0.68) & 0.73 (0.59 - 0.83) \\
 & penult. & 0.74 (0.59 - 0.84) & 0.95 (0.90 - 0.97) & 0.79 (0.66 - 0.87) & 0.94 (0.90 - 0.97) \\
 \hline
    \end{tabular}
    \caption{95\% confidence interval for Spearman rank correlation between weak alignment and CTL, obtained with BCA bootstraping with 2000 resamples.}
    \label{tab:confidence_weak}
\end{table*}

\begin{table*}[t]
    \centering
    \begin{tabular}{l|l|cc|cc}
task & layer & \multicolumn{2}{c|}{en-X} & \multicolumn{2}{c}{X-en}\\
& & before & after & before & after \\
\hline
\multirow{2}{*}{POS} & last & 0.80 (0.73 - 0.86) & 0.85 (0.81 - 0.88) & 0.83 (0.77 - 0.87) & 0.87 (0.83 - 0.91) \\
 & penult. & 0.86 (0.79 - 0.89) & 0.85 (0.79 - 0.89) & 0.87 (0.82 - 0.91) & 0.86 (0.82 - 0.90) \\
 \hline
\multirow{2}{*}{NER} & last & 0.85 (0.78 - 0.90) & 0.66 (0.53 - 0.77) & 0.86 (0.82 - 0.90) & 0.74 (0.65 - 0.82) \\
 & penult. & 0.87 (0.83 - 0.91) & 0.75 (0.65 - 0.84) & 0.88 (0.84 - 0.92) & 0.76 (0.66 - 0.84) \\
 \hline
 \multirow{2}{*}{NLI} & last & 0.74 (0.63 - 0.82) & 0.81 (0.72 - 0.87) & 0.89 (0.83 - 0.93) & 0.84 (0.77 - 0.90) \\
 & penult. & 0.84 (0.79 - 0.88) & 0.92 (0.87 - 0.95) & 0.90 (0.86 - 0.93) & 0.94 (0.90 - 0.96) \\
 \hline
    \end{tabular}
    \caption{95\% confidence interval for Spearman rank correlation between strong alignment and cross-lingual transfer, obtained with BCA bootstraping with 2000 resamples.}
    \label{tab:confidence_strong}
\end{table*}

\section{Confidence intervals for correlation}\label{section:ci_correlation}

In Section \ref{section:correlation} we compared correlation for different tasks, before and after fine-tuning, for English-target and target-English alignment and for the last and penultimate layer. These correlations where computed across several models, languages and seeds. From this correlation statistics, we have drawn three conclusions:

\begin{enumerate}
    \item Strong alignment is better correlated with cross-lingual transfer than weak alignment.
    \item The NLI task, because of its sentence-level nature, have a cross-lingual transfer that correlates better with the penultimate layer than the last one.
    \item The results do not significantly attribute higher correlation of cross-lingual transfer with alignment before or after fine-tuning neither with English-target compared to target-English alignment. 
\end{enumerate}

We verify here that these conclusions hold when looking at the confidence intervals (Tab. \ref{tab:confidence_weak} and Tab. \ref{tab:confidence_strong}). Confidence intervals are obtained using the Bias-Corrected and Accelerated (BCA) bootstrap method, where several subsets (2000) subsets of our 100 points for each measure of the correlation coefficient are sampled to obtain an empirical distribution of the correlation from which the confidence interval can be deduced \cite{Efron1994AnIT}. Since we are dealing with ordinal data (the rank in Spearman's rank correlation), bootstrap confidence intervals are expected to have better properties than methods based on assumptions about the distribution \cite{Ruscio2008ConstructingCI,Bishara2017}

Is strong alignment significantly better correlated with cross-lingual transfer than weak alignment? comparing both tables cell-by-cell reveals that confidence intervals for the last layer before fine-tuning hardly never overlap, and when they do it's with a small overlap. So in the case of alignment of the last layer before fine-tuning, strong alignment is significantly better correlated with cross-lingual transfer than weak alignment. For other situations, confidence interval overlap. But the fact that strong alignment has almost systematically a higher correlation makes our correlation still relevant.

Does the penultimate layer correlate better than the last one for NLI? For this task, we observe that the confidence intervals of the penultimate and last layer do not overlap when the alignment is measured after fine-tuning. Otherwise, before fine-tuning, we can still observe that the measured correlation for the penultimate layer is systematically above the confidence interval for the last layer, except for target-English strong alignment. 

We can see that confidence intervals overlap too much when comparing before and after fine-tuning, except in two cases. When looking at POS-tagging for the last layer, weak alignment after fine-tuning gives a significantly better correlation than before, but this does not translate to strong alignment which correlates better with cross-lingual transfer overall. The same observation can be made about NLI for the penultimate layer. On the other hand, for the NER task, strong alignment after-fine tuning gives a significantly worse correlation than before. It is thus difficult to conclude on whether alignment before or after fine-tuning is better correlated to cross-lingual transfer.

Finally, comparing target-English and English-target alignment does not give significant results. If all other parameters are kept identical, every situation leads to an overlap between confidence intervals except for the last layer before fine-tuning for NLI, which might just be fortuitous since it's the second before last layer that correlates better with cross-lingual transfer for this task. 

\begin{table*}[t]
    \centering
    \begin{tabular}{l|l|c|c|c|c|c}
task & model & en-ar & en-es & en-fr & en-ru & en-zh\\                                                                                                                                                                                                                          
                                                                                                                          
\hline\multirow{4}{*}{POS}& distilmBERT &  -0.74$_{\pm 0.11}$ &  -0.86$_{\pm 0.04}$ &  -0.87$_{\pm 0.03}$ &  -0.87$_{\pm 0.01}$ &  -0.96$_{\pm 0.04}$\\                                                                                                                         
& mBERT &  -0.90$_{\pm 0.04}$ &  -0.86$_{\pm 0.04}$ &  -0.93$_{\pm 0.02}$ &  -0.95$_{\pm 0.01}$ &  -0.96$_{\pm 0.02}$\\                                                                                                                                                         
& XLM-R Base &  -0.43$_{\pm 0.18}$ &  -0.46$_{\pm 0.10}$ &  -0.46$_{\pm 0.17}$ &  -0.70$_{\pm 0.05}$ &  0.69$_{\pm 0.40}$\\                                                                                                                                                     
& XLM-R Large &  -0.30$_{\pm 0.17}$ &  0.23$_{\pm 0.28}$ &  0.44$_{\pm 0.30}$ &  -0.44$_{\pm 0.14}$ &  0.26$_{\pm 0.25}$\\                                                                                                                                                      
\hline\multirow{4}{*}{NER}& distilmBERT &  0.00$_{\pm 0.37}$ &  -0.61$_{\pm 0.05}$ &  -0.60$_{\pm 0.05}$ &  -0.33$_{\pm 0.09}$ &  0.00$_{\pm 0.22}$\\                                                                                                                           
& mBERT &  -0.28$_{\pm 0.19}$ &  -0.36$_{\pm 0.12}$ &  -0.49$_{\pm 0.11}$ &  -0.27$_{\pm 0.20}$ &  -0.25$_{\pm 0.13}$\\                                                                                                                                                         
& XLM-R Base &  5.88$_{\pm 2.69}$ &  0.22$_{\pm 0.29}$ &  0.62$_{\pm 0.47}$ &  1.32$_{\pm 0.50}$ &  21.99$_{\pm 6.44}$\\
& XLM-R Large &  16.34$_{\pm 8.76}$ &  2.22$_{\pm 0.44}$ &  3.17$_{\pm 0.89}$ &  3.10$_{\pm 0.72}$ &  12.67$_{\pm 3.97}$\\
\hline\multirow{4}{*}{NLI}& distilmBERT &  5.49$_{\pm 0.69}$ &  0.30$_{\pm 0.11}$ &  0.88$_{\pm 0.14}$ &  2.28$_{\pm 0.36}$ &  9.78$_{\pm 0.90}$\\                                                                                                                              
& mBERT &  5.65$_{\pm 1.45}$ &  0.99$_{\pm 0.70}$ &  1.08$_{\pm 0.63}$ &  1.45$_{\pm 0.63}$ &  5.85$_{\pm 0.62}$\\                                                                                                                                                              
& XLM-R Base &  11.17$_{\pm 0.95}$ &  1.01$_{\pm 0.12}$ &  1.55$_{\pm 0.28}$ &  2.67$_{\pm 0.24}$ &  27.58$_{\pm 3.33}$\\                                                                                                                                                       
& XLM-R Large &  25.36$_{\pm 10.33}$ &  1.78$_{\pm 0.77}$ &  2.96$_{\pm 1.18}$ &  2.99$_{\pm 1.15}$ &  13.57$_{\pm 5.86}$\\                           
\hline

    \end{tabular}
    \caption{Relative variation of strong alignment at the last layer before and after fine-tuning for different fine-tuning tasks. "$_{\pm}$" indicates standard deviation. }
    \label{tab:drop_alignment_detailed} %TODO add complete results in Appendix
\end{table*}

\section{Detailed results for alignment drop}
\label{section:detailed_drop}

Tab. \ref{tab:drop_alignment_detailed} contains the detailed results when measuring the relative drop in strong alignment after fine-tuning. This is a detailed version of Tab. \ref{tab:drop_alignment} in Section \ref{section:fine-tuning}, with standard deviation measured over 5 different seeds for model initialization and data shuffling for fine-tuning, and all tested languages. This confirms that the observed increases and decreases in alignment are significant. It also seems to show that alignment for distant languages (en-ar, en-zh) is more affected by fine-tuning than other pairs.

\section{Breaking down correlation by models and layers}\label{section:correlation_breakdown}

\begin{table}[]
    \centering
    \begin{tabular}{c|cc}
 & before & after\\
\hline
last& 0.89 (0.64 - 0.96) & 0.83 (0.68 - 0.94)\\
-1& 0.79 (0.63 - 0.89) & 0.79 (0.64 - 0.88)\\
-2 & 0.79 (0.65 - 0.89) & 0.78 (0.65 - 0.89)\\
-3& 0.79 (0.64 - 0.89) & 0.82 (0.69 - 0.91)\\
-4& 0.79 (0.65 - 0.89) & 0.76 (0.62 - 0.86)\\
-5 & 0.79 (0.64 - 0.89) & 0.79 (0.64 - 0.91)\\
-6& 0.79 (0.66 - 0.90) & 0.77 (0.62 - 0.87)\\
    \end{tabular}
    \caption{Correlation between strong English-target alignment and CTL from English to target language for the POS-tagging task, with 95\% confidence intervals}
    \label{tab:drop_alignment_breakdown_ci} %TODO add complete results in Appendix
\end{table}

Tab. \ref{tab:drop_alignment_breakdown} shows a breakdown of the correlation between strong alignment and CTL across layers and models. These results tend to show that smaller models (distilmBERT and mBERT) have a better correlation at the last layer than larger models. It is also interesting to note that several correlation values are identical for alignment before fine-tuning, this might be explained by the fact that the seed of fine-tuning has unsurprisingly no effect on alignment measured before fine-tuning and by the possibility that alignment measured at one layer might be almost perfectly correlated with alignment at another, especially when the correlation is measured across few languages.

\begin{table*}[t]
    \centering
    \begin{tabular}{c|cc|cc|cc|cc}
& \multicolumn{2}{c|}{distilmBERT} & \multicolumn{2}{c|}{mBERT} & \multicolumn{2}{c|}{XLM-R Base} & \multicolumn{2}{c}{XLM-R Large}\\
 & before & after & before & after & before & after & before & after\\
\hline
last & 0.89 &  0.83 & 0.89 &  0.82 & 0.59 &  0.67 & 0.75 &  0.78\\
-1 & 0.79 &  0.79 & 0.80 &  0.88 & 0.59 &  0.68 & 0.75 &  0.75 \\
-2 & 0.79 &  0.78 & 0.80 &  0.84 & 0.59 &  0.68 & 0.75 &  0.76  \\
-3 & 0.79 &  0.82 & 0.89 &  0.86 & 0.59 &  0.69 & 0.75 &  0.76 \\
-4 & 0.79 &  0.76 & 0.89 &  0.83 & 0.59 &  0.68 & 0.75 &  0.74 \\
-5 & 0.79 &  0.79 & 0.80 &  0.81 & 0.59 &  0.65 & 0.65 &  0.72 \\
-6 & 0.79 &  0.77 & 0.80 &  0.80 & 0.69 &  0.64 & 0.65 &  0.70 \\
-7 & - & - & 0.80 &  0.77 & 0.69 &  0.66 & 0.65 &  0.65 \\
-8 & - & - & 0.80 &  0.77 & 0.69 &  0.66 & 0.65 &  0.62 \\
-9 & - & - & 0.80 &  0.77 & 0.69 &  0.66 & 0.65 &  0.65 \\
-10 & - & - & 0.80 &  0.79 & 0.69 &  0.68 & 0.65 &  0.69 \\
-11 & - & - & 0.80 &  0.80 & 0.69 &  0.66 & 0.65 &  0.67 \\
-12 & - & - & 0.89 &  0.89 & 0.69 &  0.68 & 0.65 &  0.67 \\
-13 & - & - & - & - & - & - & 0.75 &  0.68 \\
-14 & - & - & - & - & - & - & 0.75 &  0.76 \\
-15 & - & - & - & - & - & - & 0.75 &  0.76 \\
-16 & - & - & - & - & - & - & 0.75 &  0.76 \\
-17 & - & - & - & - & - & - & 0.75 &  0.71 \\
-18 & - & - & - & - & - & - & 0.75 &  0.72 \\
-19 & - & - & - & - & - & - & 0.75 &  0.72 \\
-20 & - & - & - & - & - & - & 0.75 &  0.73 \\
-21 & - & - & - & - & - & - & 0.75 &  0.70 \\
-22 & - & - & - & - & - & - & 0.75 &  0.70 \\
-23 & - & - & - & - & - & - & 0.75 &  0.71 \\
-24 & - & - & - & - & - & - & 0.75 &  0.75 \\
    \end{tabular}
    \caption{Correlation between strong English-target alignment and CTL from English to target language for the POS-tagging task. -i indicate depth of the model, with -1 being the second-before-last layer, and -2 the third-before-last, etc...}
    \label{tab:drop_alignment_breakdown} %TODO add complete results in Appendix
\end{table*}

However, drawing any conclusion from those figures might be irrelevant. By breaking down results by model, we measure correlation only from 25 samples, with five languages and five seeds. Furthermore, those latter seeds have no effect on alignment measured before. Tab. \ref{tab:drop_alignment_breakdown_ci} shows a focus on distilmBERT for the same results with confidence intervals obtained with BCA bootstrapping. It demonstrates that the measured correlation is not precise enough to draw any conclusion on which layer has an alignment that is better correlated with CTL, or to determine whether alignment before or after fine-tuning is more relevant to CTL abilities. As a matter of fact, the results are so inconclusive that almost all correlation values in Tab. \ref{tab:drop_alignment_breakdown} lie in any of the confidence intervals in Tab. \ref{tab:drop_alignment_breakdown_ci}.

\section{Detailed results of the controlled experiment}
\label{section:detailed_realignment}

This section provides detailed results of realignment methods for POS-tagging and NER, for all tested models, languages, translation datasets, and methods of extraction for realignment data. It also contains results for XNLI, for which only one translation dataset (opus100) and one extraction method (bilingual dictionaries) were tested. Results are shown on Tab. \ref{tab:pos_opus} (POS, opus100), \ref{tab:pos_un} (POS, multi-UN), \ref{tab:ner_opus} (NER, opus100), \ref{tab:ner_un} (NER, multi-UN), \ref{tab:nli} (NLI, opus100).

A light gray cell indicates that the realignment method obtained an average score that is closer to the baseline with the same model than the standard deviation of the said baseline. A dark gray cell indicates that the realignment method provokes a decrease w.r.t. the baseline that is bigger than the standard deviation.

Those detailed results emphasize on the conclusions of Section \ref{section:realignment}. Using bilingual dictionaries seems to provide significant improvements more often than other methods to extract realignment pairs of words. This is particularly visible for the POS-tagging tasks, where realigning with a bilingual dictionary, with joint or sequential realignment, provides the best results. For the NER task, this is less visible, but we've already seen that, on average, bilingual dictionaries give better results (Tab. \ref{tab:controlled_lang_and_models}).

The detailed results also confirm that for smaller models and certain tasks like POS-tagging, realignment methods work better. Realignment methods for POS-tagging on distilmBERT bring significant improvement for all languages. When using a bilingual dictionary, it also brings a systematically significant improvement over the baseline for mBERT on POS-tagging. For NER, the improvement is less often significant, but realignment methods still obtain some significant improvements for some languages like Arabic. For NLI, the only model on which there are some significant improvements for some languages is distilmBERT. 

Using a supposedly higher quality translation dataset like multi-UN does not provide improvement over using opus100, which is said to be better reflecting the average quality of translation datasets \cite{wu-dredze-2020-explicit}. It might even seem that using multi-UN provide slightly worse results than opus100. There are more cases of unsignificant increase of results for multi-UN for POS-tagging and NER and also more cases of apparently significant degradation of results with respect to the baseline. This might be explained by the fact that multi-UN is a corpus obtained from translation of documents in the United Nations, which might lack diversity in their content.

Finally, we observe that realignment methods, at least with the small number of realignment steps we performed here, do not impact the evaluation on the fine-tuning language (English). Indeed, even if they sometimes provoke a decrease, namely on POS-tagging, this decrease is small, rarely of more than 0.1 points.

\begin{table*}
    \centering
    \begin{tabular}{l|c|c|c|c|c|c}
       & en & ar & es & fr & ru & zh\\                                                                                                                                                                                                                                           
\hline                                                                                                                                                                                                                                                                          
distilmBERT & \textit{\textbf{96.1}}$_{\pm 0.1}$ & \textit{51.0}$_{\pm 1.3}$ & \textit{84.1}$_{\pm 0.8}$ & \textit{85.3}$_{\pm 0.2}$ & \textit{81.2}$_{\pm 0.7}$ & \textit{64.1}$_{\pm 1.5}$ \\                                                                                 
+ before fastalign & \cellcolor{gray!30}96.1$_{\pm 0.0}$ & 63.4$_{\pm 0.5}$ & 85.6$_{\pm 0.1}$ & 86.5$_{\pm 0.1}$ & 83.7$_{\pm 0.5}$ & 66.3$_{\pm 0.5}$ \\                                                                                                                      
+ before awesome & \cellcolor{gray!30}96.1$_{\pm 0.1}$ & 63.3$_{\pm 0.9}$ & 85.4$_{\pm 0.2}$ & 86.3$_{\pm 0.1}$ & 82.9$_{\pm 0.4}$ & 66.1$_{\pm 0.5}$ \\                                                                                                                        
+ before dico & \cellcolor{gray!30}96.1$_{\pm 0.1}$ & 65.5$_{\pm 0.5}$ & 85.8$_{\pm 0.2}$ & \textbf{86.5}$_{\pm 0.2}$ & \textbf{84.7}$_{\pm 0.3}$ & \textbf{67.4}$_{\pm 0.7}$ \\                                                                                                
+ joint fastalign & \cellcolor{gray!80}96.0$_{\pm 0.1}$ & 62.9$_{\pm 0.9}$ & 85.5$_{\pm 0.2}$ & 86.2$_{\pm 0.2}$ & 82.0$_{\pm 0.5}$ & \cellcolor{gray!30}65.0$_{\pm 0.6}$ \\                                                                                                    
+ joint awesome & \cellcolor{gray!30}96.1$_{\pm 0.1}$ & 63.4$_{\pm 0.3}$ & 85.4$_{\pm 0.1}$ & 86.4$_{\pm 0.1}$ & 82.1$_{\pm 0.5}$ & \cellcolor{gray!30}64.8$_{\pm 0.5}$ \\                                                                                                      
+ joint dico & \cellcolor{gray!30}96.1$_{\pm 0.0}$ & \textbf{66.8}$_{\pm 0.6}$ & \textbf{85.8}$_{\pm 0.2}$ & 86.5$_{\pm 0.2}$ & 84.1$_{\pm 0.6}$ & 66.4$_{\pm 0.7}$ \\                                                                                                          
\hline                                                                                                                                                                                                                                                                          
mBERT & \textit{\textbf{96.7}}$_{\pm 0.0}$ & \textit{51.7}$_{\pm 1.0}$ & \textit{85.6}$_{\pm 0.3}$ & \textit{86.0}$_{\pm 0.5}$ & \textit{82.1}$_{\pm 0.7}$ & \textit{66.0}$_{\pm 0.8}$ \\                                                                                       
+ before fastalign & \cellcolor{gray!80}96.6$_{\pm 0.1}$ & 64.2$_{\pm 1.2}$ & \cellcolor{gray!30}85.8$_{\pm 0.2}$ & 86.5$_{\pm 0.3}$ & 83.9$_{\pm 0.8}$ & 67.3$_{\pm 0.9}$ \\                                                                                                   
+ before awesome & \cellcolor{gray!80}96.6$_{\pm 0.1}$ & 64.0$_{\pm 1.2}$ & 85.9$_{\pm 0.3}$ & \cellcolor{gray!30}86.5$_{\pm 0.4}$ & 83.4$_{\pm 1.0}$ & \cellcolor{gray!30}66.7$_{\pm 0.9}$ \\                                                                                  
+ before dico & \cellcolor{gray!80}96.6$_{\pm 0.1}$ & 65.1$_{\pm 0.6}$ & \textbf{86.2}$_{\pm 0.2}$ & \textbf{86.9}$_{\pm 0.3}$ & \textbf{84.4}$_{\pm 0.3}$ & \textbf{68.3}$_{\pm 0.6}$ \\                                                                                       
+ joint fastalign & \cellcolor{gray!80}96.6$_{\pm 0.0}$ & 62.8$_{\pm 1.7}$ & \cellcolor{gray!30}85.3$_{\pm 0.3}$ & \cellcolor{gray!30}86.5$_{\pm 0.3}$ & \cellcolor{gray!80}81.4$_{\pm 0.4}$ & \cellcolor{gray!30}65.2$_{\pm 0.4}$ \\                                           
+ joint awesome & \cellcolor{gray!80}96.6$_{\pm 0.1}$ & 63.3$_{\pm 1.3}$ & \cellcolor{gray!30}85.3$_{\pm 0.2}$ & \cellcolor{gray!30}86.4$_{\pm 0.5}$ & \cellcolor{gray!30}81.6$_{\pm 0.5}$ & \cellcolor{gray!30}65.7$_{\pm 0.7}$ \\                                             
+ joint dico & \cellcolor{gray!80}96.6$_{\pm 0.1}$ & \textbf{66.5}$_{\pm 0.8}$ & 86.1$_{\pm 0.2}$ & 86.9$_{\pm 0.3}$ & 83.9$_{\pm 0.6}$ & 68.1$_{\pm 0.8}$ \\                                                                                                                   
\hline                                                                                                                                                                                                                                                                          
XLM-R base & \textit{95.9}$_{\pm 0.1}$ & \textit{62.5}$_{\pm 1.3}$ & \textit{86.6}$_{\pm 0.3}$ & \textit{86.9}$_{\pm 0.1}$ & \textit{\textbf{86.9}}$_{\pm 0.6}$ & \textit{70.9}$_{\pm 0.6}$ \\                                                                                  
+ before fastalign & \cellcolor{gray!30}95.9$_{\pm 0.1}$ & 64.2$_{\pm 0.7}$ & \cellcolor{gray!30}86.7$_{\pm 0.1}$ & 87.3$_{\pm 0.1}$ & \cellcolor{gray!80}86.2$_{\pm 0.6}$ & \cellcolor{gray!80}68.5$_{\pm 0.7}$ \\                                                             
+ before awesome & \textbf{96.0}$_{\pm 0.1}$ & 64.9$_{\pm 1.5}$ & \cellcolor{gray!30}86.8$_{\pm 0.1}$ & 87.2$_{\pm 0.1}$ & \cellcolor{gray!30}86.3$_{\pm 0.7}$ & \cellcolor{gray!80}68.0$_{\pm 0.7}$ \\                                                                         
+ before dico & \cellcolor{gray!30}96.0$_{\pm 0.1}$ & \textbf{67.3}$_{\pm 1.5}$ & \cellcolor{gray!30}\textbf{86.9}$_{\pm 0.2}$ & \textbf{87.3}$_{\pm 0.1}$ & \cellcolor{gray!30}86.8$_{\pm 0.7}$ & \cellcolor{gray!30}\textbf{71.2}$_{\pm 0.6}$ \\                              
+ joint fastalign & \cellcolor{gray!30}95.9$_{\pm 0.1}$ & \cellcolor{gray!30}63.5$_{\pm 0.4}$ & \cellcolor{gray!80}86.0$_{\pm 0.2}$ & \cellcolor{gray!80}86.6$_{\pm 0.2}$ & \cellcolor{gray!80}84.6$_{\pm 0.2}$ & \cellcolor{gray!80}69.1$_{\pm 0.5}$ \\                        
+ joint awesome & \cellcolor{gray!30}96.0$_{\pm 0.1}$ & \cellcolor{gray!30}63.1$_{\pm 0.7}$ & \cellcolor{gray!80}85.8$_{\pm 0.1}$ & \cellcolor{gray!80}86.4$_{\pm 0.1}$ & \cellcolor{gray!80}84.5$_{\pm 0.3}$ & \cellcolor{gray!80}69.2$_{\pm 0.2}$ \\
+ joint dico & \cellcolor{gray!30}95.9$_{\pm 0.1}$ & 66.6$_{\pm 0.4}$ & \cellcolor{gray!30}86.6$_{\pm 0.2}$ & 87.2$_{\pm 0.1}$ & \cellcolor{gray!80}86.0$_{\pm 0.3}$ & \cellcolor{gray!30}70.6$_{\pm 0.2}$ \\
\hline
XLM-R large & \textit{\textbf{97.7}}$_{\pm 0.0}$ & \textit{\textbf{65.1}}$_{\pm 0.6}$ & \textit{\textbf{87.0}}$_{\pm 0.6}$ & \textit{\textbf{87.5}}$_{\pm 0.6}$ & \textit{\textbf{87.0}}$_{\pm 0.9}$ & \textit{\textbf{71.5}}$_{\pm 0.2}$ \\
\hline

    \end{tabular}
    \caption{Controlled experiment with realignment for POS-tagging with opus100 translation dataset.}
    \label{tab:pos_opus}
\end{table*}    

\begin{table*}
    \centering
    \begin{tabular}{l|c|c|c|c|c|c}
       & en & ar & es & fr & ru & zh\\                                                                                                                                                                                                                                           
\hline                                                                                                                                                                                                                                                                          
distilmBERT & \textit{96.1}$_{\pm 0.1}$ & \textit{51.0}$_{\pm 1.3}$ & \textit{84.1}$_{\pm 0.8}$ & \textit{85.3}$_{\pm 0.2}$ & \textit{81.2}$_{\pm 0.7}$ & \textit{64.1}$_{\pm 1.5}$ \\                                                                                          
+ before fastalign & \cellcolor{gray!30}96.0$_{\pm 0.1}$ & 61.8$_{\pm 0.6}$ & 85.2$_{\pm 0.2}$ & 86.1$_{\pm 0.2}$ & 82.0$_{\pm 0.4}$ & 65.7$_{\pm 0.7}$ \\                                                                                                                      
+ before awesome & \cellcolor{gray!30}96.0$_{\pm 0.1}$ & 62.1$_{\pm 0.4}$ & 85.3$_{\pm 0.2}$ & 86.1$_{\pm 0.4}$ & 82.2$_{\pm 0.5}$ & \cellcolor{gray!30}65.3$_{\pm 0.5}$ \\                                                                                                     
+ before dico & \cellcolor{gray!30}96.1$_{\pm 0.0}$ & 64.1$_{\pm 0.9}$ & \textbf{85.8}$_{\pm 0.1}$ & 86.5$_{\pm 0.1}$ & \textbf{84.6}$_{\pm 0.6}$ & \textbf{67.4}$_{\pm 1.0}$ \\                                                                                                
+ joint fastalign & \cellcolor{gray!30}96.1$_{\pm 0.1}$ & 62.8$_{\pm 0.6}$ & 85.2$_{\pm 0.1}$ & 86.1$_{\pm 0.1}$ & \cellcolor{gray!30}81.2$_{\pm 0.4}$ & \cellcolor{gray!30}64.7$_{\pm 0.6}$ \\                                                                                 
+ joint awesome & \cellcolor{gray!30}\textbf{96.1}$_{\pm 0.1}$ & 61.8$_{\pm 0.8}$ & 85.2$_{\pm 0.2}$ & 86.0$_{\pm 0.2}$ & \cellcolor{gray!30}81.2$_{\pm 0.4}$ & \cellcolor{gray!30}64.6$_{\pm 0.5}$ \\                                                                          
+ joint dico & \cellcolor{gray!30}96.1$_{\pm 0.0}$ & \textbf{65.5}$_{\pm 0.4}$ & 85.8$_{\pm 0.2}$ & \textbf{86.7}$_{\pm 0.1}$ & 83.7$_{\pm 0.6}$ & 65.9$_{\pm 0.8}$ \\                                                                                                          
\hline                                                                                                                                                                                                                                                                          
mBERT & \textit{96.7}$_{\pm 0.0}$ & \textit{51.7}$_{\pm 1.0}$ & \textit{85.6}$_{\pm 0.3}$ & \textit{86.0}$_{\pm 0.5}$ & \textit{82.1}$_{\pm 0.7}$ & \textit{66.0}$_{\pm 0.8}$ \\                                                                                                
+ before fastalign & \cellcolor{gray!80}96.6$_{\pm 0.0}$ & 63.1$_{\pm 0.6}$ & \cellcolor{gray!30}85.6$_{\pm 0.1}$ & \cellcolor{gray!30}86.4$_{\pm 0.2}$ & \cellcolor{gray!30}82.7$_{\pm 0.6}$ & 66.9$_{\pm 0.9}$ \\                                                             
+ before awesome & \textbf{96.7}$_{\pm 0.1}$ & 61.8$_{\pm 1.0}$ & \cellcolor{gray!30}85.6$_{\pm 0.3}$ & \cellcolor{gray!30}86.5$_{\pm 0.3}$ & \cellcolor{gray!30}82.1$_{\pm 0.8}$ & \cellcolor{gray!30}66.6$_{\pm 0.8}$ \\                                                      
+ before dico & \cellcolor{gray!80}96.6$_{\pm 0.1}$ & 64.2$_{\pm 1.4}$ & \textbf{86.0}$_{\pm 0.3}$ & 86.9$_{\pm 0.4}$ & \textbf{84.1}$_{\pm 0.8}$ & \textbf{69.0}$_{\pm 0.8}$ \\                                                                                                
+ joint fastalign & \cellcolor{gray!30}96.7$_{\pm 0.0}$ & 62.6$_{\pm 1.0}$ & \cellcolor{gray!30}85.4$_{\pm 0.3}$ & \cellcolor{gray!30}86.3$_{\pm 0.5}$ & \cellcolor{gray!80}80.7$_{\pm 0.8}$ & \cellcolor{gray!80}65.0$_{\pm 0.6}$ \\                                           
+ joint awesome & \cellcolor{gray!30}96.7$_{\pm 0.0}$ & 61.9$_{\pm 0.8}$ & \cellcolor{gray!80}85.2$_{\pm 0.3}$ & \cellcolor{gray!30}86.1$_{\pm 0.3}$ & \cellcolor{gray!80}80.9$_{\pm 0.4}$ & \cellcolor{gray!80}64.9$_{\pm 0.6}$ \\                                             
+ joint dico & \cellcolor{gray!80}96.6$_{\pm 0.1}$ & \textbf{64.5}$_{\pm 1.4}$ & 86.0$_{\pm 0.4}$ & \textbf{86.9}$_{\pm 0.5}$ & 83.9$_{\pm 0.9}$ & 67.3$_{\pm 0.7}$ \\                                                                                                          
\hline                                                                                                                                                                                                                                                                          
XLM-R base & \textit{95.9}$_{\pm 0.1}$ & \textit{62.5}$_{\pm 1.3}$ & \textit{86.6}$_{\pm 0.3}$ & \textit{86.9}$_{\pm 0.1}$ & \textit{\textbf{86.9}}$_{\pm 0.6}$ & \textit{70.9}$_{\pm 0.6}$ \\
+ before fastalign & \cellcolor{gray!30}95.9$_{\pm 0.1}$ & 64.0$_{\pm 1.0}$ & \cellcolor{gray!80}86.3$_{\pm 0.1}$ & \cellcolor{gray!30}87.0$_{\pm 0.3}$ & \cellcolor{gray!80}85.8$_{\pm 0.7}$ & \cellcolor{gray!80}68.6$_{\pm 0.9}$ \\
+ before awesome & \textbf{96.0}$_{\pm 0.1}$ & 64.7$_{\pm 0.9}$ & \cellcolor{gray!30}86.4$_{\pm 0.1}$ & \cellcolor{gray!80}86.8$_{\pm 0.2}$ & \cellcolor{gray!80}85.8$_{\pm 0.4}$ & \cellcolor{gray!80}66.8$_{\pm 0.0}$ \\
+ before dico & \cellcolor{gray!30}96.0$_{\pm 0.1}$ & \textbf{66.5}$_{\pm 1.2}$ & \cellcolor{gray!30}\textbf{86.8}$_{\pm 0.2}$ & \textbf{87.2}$_{\pm 0.2}$ & \cellcolor{gray!30}86.3$_{\pm 0.6}$ & \cellcolor{gray!30}\textbf{71.1}$_{\pm 0.6}$ \\
+ joint fastalign & \cellcolor{gray!80}95.9$_{\pm 0.1}$ & \cellcolor{gray!30}62.5$_{\pm 1.1}$ & \cellcolor{gray!80}85.7$_{\pm 0.2}$ & \cellcolor{gray!80}86.3$_{\pm 0.2}$ & \cellcolor{gray!80}84.1$_{\pm 0.4}$ & \cellcolor{gray!80}69.1$_{\pm 0.3}$ \\
+ joint awesome & \cellcolor{gray!30}95.9$_{\pm 0.1}$ & \cellcolor{gray!30}62.1$_{\pm 0.9}$ & \cellcolor{gray!80}85.5$_{\pm 0.1}$ & \cellcolor{gray!80}86.2$_{\pm 0.2}$ & \cellcolor{gray!80}83.9$_{\pm 0.3}$ & \cellcolor{gray!80}69.0$_{\pm 0.4}$ \\
+ joint dico & \cellcolor{gray!80}95.9$_{\pm 0.1}$ & 65.8$_{\pm 0.7}$ & \cellcolor{gray!30}86.4$_{\pm 0.3}$ & 87.1$_{\pm 0.1}$ & \cellcolor{gray!80}85.3$_{\pm 0.6}$ & \cellcolor{gray!30}70.5$_{\pm 0.2}$ \\
\hline
XLM-R large & \textit{\textbf{97.7}}$_{\pm 0.0}$ & \textit{\textbf{65.1}}$_{\pm 0.6}$ & \textit{\textbf{87.0}}$_{\pm 0.6}$ & \textit{\textbf{87.5}}$_{\pm 0.6}$ & \textit{\textbf{87.0}}$_{\pm 0.9}$ & \textit{\textbf{71.5}}$_{\pm 0.2}$ \\
\hline
    \end{tabular}
    \caption{Controlled experiment with realignment for POS-tagging with multiUN translation dataset.}
    \label{tab:pos_un}
\end{table*}    

\begin{table*}
    \centering
    \begin{tabular}{l|c|c|c|c|c|c}
         & en & ar & es & fr & ru & zh\\
\hline
distilmBERT & \textit{82.9}$_{\pm 0.4}$ & \textit{34.5}$_{\pm 1.6}$ & \textit{69.2}$_{\pm 3.1}$ & \textit{76.1}$_{\pm 0.7}$ & \textit{60.2}$_{\pm 0.9}$ & \textit{46.8}$_{\pm 1.9}$ \\
+ before fastalign & \cellcolor{gray!30}82.9$_{\pm 0.3}$ & 39.1$_{\pm 2.0}$ & \cellcolor{gray!30}70.1$_{\pm 2.7}$ & \cellcolor{gray!30}75.9$_{\pm 0.3}$ & \cellcolor{gray!30}60.1$_{\pm 0.8}$ & \cellcolor{gray!30}46.5$_{\pm 2.5}$ \\
+ before awesome & \cellcolor{gray!30}82.7$_{\pm 0.3}$ & 39.3$_{\pm 3.7}$ & 72.5$_{\pm 2.5}$ & \cellcolor{gray!30}75.6$_{\pm 0.5}$ & \cellcolor{gray!30}60.3$_{\pm 0.3}$ & \textbf{49.7}$_{\pm 1.1}$ \\
+ before dico & \cellcolor{gray!30}82.9$_{\pm 0.4}$ & 41.6$_{\pm 1.7}$ & \cellcolor{gray!30}67.9$_{\pm 2.7}$ & \cellcolor{gray!30}76.4$_{\pm 0.9}$ & \cellcolor{gray!30}60.3$_{\pm 1.2}$ & \cellcolor{gray!30}48.3$_{\pm 1.6}$ \\
+ joint fastalign & \cellcolor{gray!30}83.0$_{\pm 0.2}$ & 41.5$_{\pm 3.1}$ & \textbf{73.5}$_{\pm 2.1}$ & \cellcolor{gray!30}76.6$_{\pm 0.7}$ & \textbf{61.4}$_{\pm 0.8}$ & \cellcolor{gray!30}48.3$_{\pm 1.5}$ \\
+ joint awesome & \cellcolor{gray!30}\textbf{83.1}$_{\pm 0.1}$ & 41.4$_{\pm 0.4}$ & \cellcolor{gray!30}72.3$_{\pm 0.1}$ & \textbf{77.6}$_{\pm 0.7}$ & \cellcolor{gray!30}60.9$_{\pm 0.4}$ & 49.5$_{\pm 0.4}$ \\
+ joint dico & \cellcolor{gray!30}83.0$_{\pm 0.5}$ & \textbf{42.2}$_{\pm 2.7}$ & \cellcolor{gray!30}69.5$_{\pm 2.3}$ & \cellcolor{gray!30}76.6$_{\pm 1.0}$ & 61.2$_{\pm 0.8}$ & 48.8$_{\pm 1.7}$ \\
\hline
mBERT & \textit{84.4}$_{\pm 0.4}$ & \textit{40.7}$_{\pm 2.9}$ & \textit{74.3}$_{\pm 1.4}$ & \textit{79.9}$_{\pm 1.3}$ & \textit{63.9}$_{\pm 2.0}$ & \textit{52.1}$_{\pm 1.7}$ \\
+ before fastalign & \cellcolor{gray!30}84.3$_{\pm 0.4}$ & \cellcolor{gray!30}42.0$_{\pm 2.9}$ & \cellcolor{gray!80}70.5$_{\pm 3.0}$ & \cellcolor{gray!30}79.3$_{\pm 0.6}$ & \cellcolor{gray!30}\textbf{65.5}$_{\pm 1.6}$ & \cellcolor{gray!30}51.7$_{\pm 1.1}$ \\
+ before awesome & \cellcolor{gray!30}\textbf{84.8}$_{\pm 0.2}$ & \cellcolor{gray!30}40.2$_{\pm 2.6}$ & \cellcolor{gray!80}72.3$_{\pm 2.8}$ & \cellcolor{gray!30}79.4$_{\pm 0.7}$ & \cellcolor{gray!30}63.0$_{\pm 1.8}$ & \cellcolor{gray!30}51.2$_{\pm 0.6}$ \\
+ before dico & \cellcolor{gray!30}84.3$_{\pm 0.6}$ & \cellcolor{gray!30}42.1$_{\pm 1.7}$ & \cellcolor{gray!30}73.4$_{\pm 2.8}$ & \cellcolor{gray!30}80.1$_{\pm 1.4}$ & \cellcolor{gray!30}64.9$_{\pm 1.7}$ & \cellcolor{gray!30}52.8$_{\pm 1.2}$ \\
+ joint fastalign & \cellcolor{gray!30}84.3$_{\pm 0.3}$ & \cellcolor{gray!30}42.7$_{\pm 1.9}$ & \cellcolor{gray!30}75.6$_{\pm 2.0}$ & \cellcolor{gray!30}80.5$_{\pm 1.0}$ & \cellcolor{gray!30}65.4$_{\pm 1.5}$ & 54.3$_{\pm 1.2}$ \\
+ joint awesome & \cellcolor{gray!30}84.1$_{\pm 0.4}$ & 44.2$_{\pm 2.2}$ & \cellcolor{gray!30}75.6$_{\pm 1.6}$ & \cellcolor{gray!30}80.2$_{\pm 0.2}$ & \cellcolor{gray!30}64.8$_{\pm 2.4}$ & 54.6$_{\pm 1.1}$ \\
+ joint dico & \cellcolor{gray!30}84.2$_{\pm 0.3}$ & \textbf{46.0}$_{\pm 3.2}$ & \textbf{76.6}$_{\pm 1.9}$ & \cellcolor{gray!30}\textbf{81.1}$_{\pm 0.9}$ & \cellcolor{gray!30}65.5$_{\pm 0.9}$ & \textbf{54.9}$_{\pm 0.9}$ \\
\hline
XLM-R base & \textit{80.0}$_{\pm 0.3}$ & \textit{46.4}$_{\pm 2.7}$ & \textit{71.8}$_{\pm 3.7}$ & \textit{75.0}$_{\pm 1.4}$ & \textit{61.6}$_{\pm 0.8}$ & \textit{47.4}$_{\pm 2.1}$ \\
+ before fastalign & \cellcolor{gray!30}\textbf{80.2}$_{\pm 0.4}$ & 51.5$_{\pm 3.1}$ & \cellcolor{gray!30}71.7$_{\pm 1.4}$ & \cellcolor{gray!30}75.9$_{\pm 1.0}$ & \cellcolor{gray!30}62.1$_{\pm 1.2}$ & \cellcolor{gray!30}45.7$_{\pm 1.3}$ \\
+ before awesome & \cellcolor{gray!30}80.1$_{\pm 0.3}$ & 52.2$_{\pm 3.4}$ & \cellcolor{gray!30}74.2$_{\pm 1.3}$ & \cellcolor{gray!30}76.1$_{\pm 0.8}$ & \cellcolor{gray!30}61.0$_{\pm 1.7}$ & \cellcolor{gray!30}46.6$_{\pm 1.0}$ \\
+ before dico & \cellcolor{gray!30}80.0$_{\pm 0.2}$ & \textbf{55.8}$_{\pm 3.6}$ & \textbf{76.9}$_{\pm 1.6}$ & \textbf{77.3}$_{\pm 0.7}$ & \cellcolor{gray!30}62.0$_{\pm 0.4}$ & \cellcolor{gray!30}47.5$_{\pm 0.7}$ \\
+ joint fastalign & \cellcolor{gray!30}79.8$_{\pm 0.2}$ & \cellcolor{gray!30}47.7$_{\pm 5.0}$ & \cellcolor{gray!30}74.2$_{\pm 1.2}$ & \cellcolor{gray!30}75.6$_{\pm 0.7}$ & 63.0$_{\pm 0.8}$ & 50.2$_{\pm 1.5}$ \\
+ joint awesome & \cellcolor{gray!30}79.7$_{\pm 0.3}$ & \cellcolor{gray!30}47.6$_{\pm 3.1}$ & \cellcolor{gray!30}73.9$_{\pm 1.2}$ & \cellcolor{gray!30}75.4$_{\pm 0.4}$ & 63.5$_{\pm 0.8}$ & \textbf{51.2}$_{\pm 1.1}$ \\
+ joint dico & \cellcolor{gray!30}79.9$_{\pm 0.3}$ & 50.3$_{\pm 3.2}$ & \cellcolor{gray!30}75.2$_{\pm 1.1}$ & \cellcolor{gray!30}75.9$_{\pm 0.8}$ & \textbf{63.6}$_{\pm 0.6}$ & 50.1$_{\pm 1.3}$ \\
\hline
XLM-R large & \textit{\textbf{83.8}}$_{\pm 1.0}$ & \textit{\textbf{45.1}}$_{\pm 1.4}$ & \textit{\textbf{75.6}}$_{\pm 3.7}$ & \textit{\textbf{80.7}}$_{\pm 0.8}$ & \textit{\textbf{70.5}}$_{\pm 3.4}$ & \textit{\textbf{53.0}}$_{\pm 2.1}$ \\
\hline
    \end{tabular}
    \caption{Controlled experiment with realignment for NER with opus100 translation dataset.}
    \label{tab:ner_opus}
\end{table*}  

\begin{table*}
    \centering
    \begin{tabular}{l|c|c|c|c|c|c}
        & same & ar & es & fr & ru & zh\\
\hline
distilmBERT & \textit{82.9}$_{\pm 0.4}$ & \textit{34.5}$_{\pm 1.6}$ & \textit{69.2}$_{\pm 3.1}$ & \textit{76.1}$_{\pm 0.7}$ & \textit{60.2}$_{\pm 0.9}$ & \textit{46.8}$_{\pm 1.9}$ \\
+ before fastalign & \cellcolor{gray!30}82.9$_{\pm 0.3}$ & 37.8$_{\pm 0.6}$ & \cellcolor{gray!30}71.1$_{\pm 3.8}$ & \cellcolor{gray!30}76.2$_{\pm 1.3}$ & \cellcolor{gray!30}59.8$_{\pm 1.7}$ & \cellcolor{gray!30}46.9$_{\pm 2.0}$ \\
+ before awesome & \cellcolor{gray!30}83.0$_{\pm 0.4}$ & 39.4$_{\pm 1.1}$ & \cellcolor{gray!30}\textbf{71.7}$_{\pm 3.6}$ & \cellcolor{gray!30}75.7$_{\pm 0.3}$ & 61.2$_{\pm 1.7}$ & \cellcolor{gray!30}46.3$_{\pm 1.3}$ \\
+ before dico & \cellcolor{gray!30}\textbf{83.0}$_{\pm 0.1}$ & \textbf{43.3}$_{\pm 2.9}$ & \cellcolor{gray!30}69.0$_{\pm 3.5}$ & \textbf{77.6}$_{\pm 1.0}$ & \cellcolor{gray!30}59.9$_{\pm 0.9}$ & \cellcolor{gray!30}47.1$_{\pm 1.5}$ \\
+ joint fastalign & \cellcolor{gray!30}83.0$_{\pm 0.2}$ & 39.5$_{\pm 0.4}$ & \cellcolor{gray!30}70.5$_{\pm 2.3}$ & \cellcolor{gray!30}76.6$_{\pm 0.5}$ & 61.3$_{\pm 1.0}$ & \cellcolor{gray!30}48.4$_{\pm 1.7}$ \\
+ joint awesome & \cellcolor{gray!30}82.9$_{\pm 0.3}$ & 38.8$_{\pm 1.8}$ & \cellcolor{gray!30}69.6$_{\pm 1.0}$ & \cellcolor{gray!30}76.7$_{\pm 0.8}$ & \cellcolor{gray!30}60.6$_{\pm 1.4}$ & \textbf{49.2}$_{\pm 1.1}$ \\
+ joint dico & \cellcolor{gray!30}83.0$_{\pm 0.3}$ & 41.9$_{\pm 1.5}$ & \cellcolor{gray!30}71.3$_{\pm 1.7}$ & 77.5$_{\pm 1.5}$ & \textbf{61.9}$_{\pm 1.4}$ & \cellcolor{gray!30}48.4$_{\pm 1.1}$ \\
\hline
mBERT & \textit{84.4}$_{\pm 0.4}$ & \textit{40.7}$_{\pm 2.9}$ & \textit{74.3}$_{\pm 1.4}$ & \textit{79.9}$_{\pm 1.3}$ & \textit{63.9}$_{\pm 2.0}$ & \textit{52.1}$_{\pm 1.7}$ \\
+ before fastalign & \cellcolor{gray!30}84.3$_{\pm 0.4}$ & \textbf{46.2}$_{\pm 3.6}$ & \cellcolor{gray!80}69.4$_{\pm 4.2}$ & \cellcolor{gray!80}78.4$_{\pm 0.8}$ & \cellcolor{gray!30}64.3$_{\pm 1.1}$ & \cellcolor{gray!30}51.2$_{\pm 1.9}$ \\
+ before dico & \cellcolor{gray!30}\textbf{84.6}$_{\pm 0.4}$ & \cellcolor{gray!30}42.5$_{\pm 2.0}$ & \cellcolor{gray!80}71.9$_{\pm 2.6}$ & \cellcolor{gray!30}79.9$_{\pm 0.8}$ & \cellcolor{gray!30}64.0$_{\pm 1.1}$ & \cellcolor{gray!30}52.1$_{\pm 1.1}$ \\
+ joint fastalign & \cellcolor{gray!30}84.1$_{\pm 0.3}$ & 46.0$_{\pm 1.0}$ & \cellcolor{gray!30}74.4$_{\pm 2.6}$ & \cellcolor{gray!30}\textbf{80.7}$_{\pm 1.0}$ & \textbf{66.6}$_{\pm 2.4}$ & \textbf{54.8}$_{\pm 0.4}$ \\
+ joint awesome & \cellcolor{gray!30}84.1$_{\pm 0.5}$ & \cellcolor{gray!30}42.2$_{\pm 2.2}$ & \cellcolor{gray!30}\textbf{74.9}$_{\pm 0.5}$ & \cellcolor{gray!30}80.3$_{\pm 0.6}$ & \cellcolor{gray!30}65.5$_{\pm 2.5}$ & 54.5$_{\pm 0.9}$ \\
+ joint dico & \cellcolor{gray!30}84.3$_{\pm 0.4}$ & \cellcolor{gray!30}43.6$_{\pm 3.2}$ & \cellcolor{gray!30}74.5$_{\pm 1.2}$ & \cellcolor{gray!30}80.4$_{\pm 0.8}$ & \cellcolor{gray!30}65.6$_{\pm 3.3}$ & \cellcolor{gray!30}53.4$_{\pm 1.5}$ \\
\hline
XLM-R base & \textit{80.0}$_{\pm 0.3}$ & \textit{46.4}$_{\pm 2.7}$ & \textit{71.8}$_{\pm 3.7}$ & \textit{75.0}$_{\pm 1.4}$ & \textit{61.6}$_{\pm 0.8}$ & \textit{47.4}$_{\pm 2.1}$ \\
+ before fastalign & \cellcolor{gray!30}\textbf{80.2}$_{\pm 0.3}$ & 53.8$_{\pm 3.4}$ & \cellcolor{gray!30}72.0$_{\pm 2.8}$ & \cellcolor{gray!30}76.0$_{\pm 1.4}$ & \cellcolor{gray!30}61.8$_{\pm 1.5}$ & \cellcolor{gray!30}45.8$_{\pm 1.3}$ \\
+ before awesome & \cellcolor{gray!30}80.2$_{\pm 0.3}$ & \textbf{54.8}$_{\pm 1.3}$ & \cellcolor{gray!30}70.4$_{\pm 2.0}$ & 76.5$_{\pm 1.4}$ & \cellcolor{gray!30}61.9$_{\pm 0.0}$ & \cellcolor{gray!30}45.5$_{\pm 0.7}$ \\
+ before dico & \cellcolor{gray!30}80.0$_{\pm 0.2}$ & 54.1$_{\pm 1.7}$ & \textbf{76.1}$_{\pm 1.4}$ & \textbf{76.5}$_{\pm 0.7}$ & \cellcolor{gray!30}61.8$_{\pm 1.0}$ & \cellcolor{gray!30}47.6$_{\pm 1.4}$ \\
+ joint fastalign & \cellcolor{gray!30}79.8$_{\pm 0.2}$ & \cellcolor{gray!30}46.8$_{\pm 3.0}$ & \cellcolor{gray!30}73.5$_{\pm 2.4}$ & \cellcolor{gray!30}75.6$_{\pm 1.0}$ & \cellcolor{gray!30}61.8$_{\pm 1.5}$ & 50.0$_{\pm 1.8}$ \\
+ joint awesome & \cellcolor{gray!30}79.8$_{\pm 0.3}$ & \cellcolor{gray!30}49.0$_{\pm 3.1}$ & 75.8$_{\pm 1.7}$ & \cellcolor{gray!30}76.1$_{\pm 1.1}$ & \textbf{63.0}$_{\pm 1.5}$ & \textbf{50.8}$_{\pm 0.8}$ \\
+ joint dico & \cellcolor{gray!30}79.8$_{\pm 0.3}$ & \cellcolor{gray!30}48.4$_{\pm 3.0}$ & \cellcolor{gray!30}74.3$_{\pm 1.6}$ & \cellcolor{gray!30}75.8$_{\pm 0.8}$ & 62.7$_{\pm 0.7}$ & 50.2$_{\pm 0.1}$ \\
\hline
XLM-R large & \textit{\textbf{83.8}}$_{\pm 1.0}$ & \textit{\textbf{45.1}}$_{\pm 1.4}$ & \textit{\textbf{75.6}}$_{\pm 3.7}$ & \textit{\textbf{80.7}}$_{\pm 0.8}$ & \textit{\textbf{70.5}}$_{\pm 3.4}$ & \textit{\textbf{53.0}}$_{\pm 2.1}$ \\
\hline
    \end{tabular}
    \caption{Controlled experiment with realignment for NER with multiUN translation dataset.}
    \label{tab:ner_un}
\end{table*}  

\begin{table*}
    \centering
    \begin{tabular}{l|c|c|c|c|c|c}
        & en & ar & es & fr & ru & zh\\
\hline
distilmBERT & \textit{76.0}$_{\pm 0.7}$ & \textit{58.2}$_{\pm 1.4}$ & \textit{68.5}$_{\pm 0.6}$ & \textit{\textbf{68.7}}$_{\pm 0.6}$ & \textit{62.3}$_{\pm 1.2}$ & \textit{63.4}$_{\pm 0.9}$ \\
+ before dico & \cellcolor{gray!30}\textbf{76.2}$_{\pm 0.6}$ & \cellcolor{gray!30}58.4$_{\pm 1.2}$ & 69.2$_{\pm 0.8}$ & \cellcolor{gray!80}68.0$_{\pm 0.8}$ & \cellcolor{gray!30}62.6$_{\pm 1.1}$ & \cellcolor{gray!30}63.1$_{\pm 0.9}$ \\
+ joint dico & \cellcolor{gray!30}76.2$_{\pm 0.7}$ & \textbf{59.8}$_{\pm 1.1}$ & \textbf{69.2}$_{\pm 1.0}$ & \cellcolor{gray!30}68.6$_{\pm 1.1}$ & \cellcolor{gray!30}\textbf{63.0}$_{\pm 1.0}$ & \textbf{64.4}$_{\pm 1.2}$ \\
\hline
mBERT & \textit{\textbf{80.2}}$_{\pm 0.7}$ & \textit{65.2}$_{\pm 1.2}$ & \textit{73.8}$_{\pm 0.8}$ & \textit{\textbf{72.9}}$_{\pm 0.7}$ & \textit{68.3}$_{\pm 1.2}$ & \textit{68.5}$_{\pm 1.3}$ \\
+ before dico & \cellcolor{gray!80}79.0$_{\pm 0.6}$ & \cellcolor{gray!80}63.2$_{\pm 0.9}$ & \cellcolor{gray!80}72.9$_{\pm 0.8}$ & \cellcolor{gray!80}71.7$_{\pm 0.5}$ & \cellcolor{gray!80}66.9$_{\pm 0.9}$ & \cellcolor{gray!30}68.7$_{\pm 0.7}$ \\
+ joint dico & \cellcolor{gray!30}79.9$_{\pm 0.9}$ & \cellcolor{gray!30}\textbf{65.6}$_{\pm 0.9}$ & \cellcolor{gray!30}\textbf{73.8}$_{\pm 1.3}$ & \cellcolor{gray!30}72.5$_{\pm 1.2}$ & \cellcolor{gray!30}\textbf{68.8}$_{\pm 1.3}$ & \cellcolor{gray!30}\textbf{69.0}$_{\pm 0.8}$ \\
\hline
XLM-R base & \textit{82.8}$_{\pm 1.6}$ & \textit{70.1}$_{\pm 1.4}$ & \textit{77.4}$_{\pm 1.6}$ & \textit{76.5}$_{\pm 1.3}$ & \textit{74.2}$_{\pm 1.3}$ & \textit{71.7}$_{\pm 1.3}$ \\
+ before dico & \cellcolor{gray!30}81.2$_{\pm 2.4}$ & \cellcolor{gray!80}68.4$_{\pm 2.9}$ & \cellcolor{gray!30}76.0$_{\pm 2.2}$ & \cellcolor{gray!80}74.9$_{\pm 1.9}$ & \cellcolor{gray!80}72.8$_{\pm 2.4}$ & \cellcolor{gray!30}71.4$_{\pm 2.5}$ \\
+ joint dico & \cellcolor{gray!30}\textbf{83.7}$_{\pm 0.7}$ & \cellcolor{gray!30}\textbf{70.8}$_{\pm 1.4}$ & \cellcolor{gray!30}\textbf{78.0}$_{\pm 0.9}$ & \cellcolor{gray!30}\textbf{76.7}$_{\pm 1.0}$ & \cellcolor{gray!30}\textbf{74.6}$_{\pm 1.3}$ & \cellcolor{gray!30}\textbf{72.7}$_{\pm 1.6}$ \\
\hline
XLM-R large & \textit{\textbf{87.9}}$_{\pm 0.7}$ & \textit{\textbf{77.5}}$_{\pm 1.3}$ & \textit{\textbf{83.2}}$_{\pm 1.3}$ & \textit{\textbf{81.9}}$_{\pm 1.2}$ & \textit{\textbf{79.1}}$_{\pm 1.1}$ & \textit{\textbf{78.2}}$_{\pm 1.3}$ \\
\hline
    \end{tabular}
    \caption{Controlled experiment with realignment for NLI with opus100 translation dataset.}
    \label{tab:nli}
\end{table*}

\end{document}